\documentclass{article}
\usepackage{arxiv}

\usepackage[utf8]{inputenc}

\usepackage[numbers]{natbib}
\usepackage{booktabs}
\usepackage{hyperref}
\usepackage{abstract}
\usepackage{authblk}

\date{\vskip -0.4in}

\usepackage{xcolor}
\usepackage{graphicx}
\usepackage{numprint}

\usepackage{bbm}
\usepackage{fancyvrb}
\usepackage{xspace}
\usepackage{pythonhighlight}
\usepackage{subcaption}
\usepackage{caption}
\usepackage{csquotes}
\usepackage{enumitem}

\usepackage{amsmath}
\usepackage{amsthm}
\usepackage{amssymb}
\usepackage{bm}

\theoremstyle{definition}
\newtheorem{definition}{Definition}

\usepackage{tikz}
\usetikzlibrary{bayesnet}
\usetikzlibrary{arrows}
\usetikzlibrary{fit}
\usetikzlibrary{shapes.geometric}
\usetikzlibrary{shapes.misc}

\usepackage{xcolor}
\definecolor{azure}{rgb}{0.0, 0.5, 1.0}

\definecolor{nvidiagreen}{rgb}{0.4627, 0.7255, 0.0}

\definecolor{darkred}{rgb}{0.7, 0, 0}
\definecolor{cadmiumorange}{rgb}{0.93, 0.53, 0.18}

\newcounter{requirementno}
\renewcommand{\therequirementno}{\Roman{requirementno}}

\newcommand{\reqentry}[1]{\refstepcounter{requirementno}{\therequirementno}\label{#1}}

\newcommand{\dppp}{\texttt{d3p}\xspace}

\usepackage{listings}
\definecolor{keywordcolor}{rgb}{0.38, 0, 1}

\usepackage[space=true]{accsupp}
\newcommand{\copyablespace}{\BeginAccSupp{method=hex,unicode,ActualText=00A0}\ \EndAccSupp{}}
\title{\huge d3p - A Python Package for Differentially-Private Probabilistic Programming}

\author[1]{Lukas Prediger}
\author[2]{Niki Loppi}
\author[1,3]{Samuel Kaski}
\author[4]{Antti Honkela}

\affil[1]{Aalto University, Finland}
\affil[2]{NVIDIA AI Technology Center, Finland}
\affil[3]{University of Manchester, UK}
\affil[4]{University of Helsinki, Finland}

\begin{document}

\maketitle

\begin{abstract}{
We present \dppp, a software package designed to help fielding runtime efficient widely-applicable Bayesian inference under differential privacy guarantees. \dppp achieves general applicability to a wide range of probabilistic modelling problems by implementing the differentially private variational inference algorithm, allowing users to fit any parametric probabilistic model with a differentiable density function. \dppp adopts the probabilistic programming paradigm as a powerful way for the user to flexibly define such models. We demonstrate the use of our software on a hierarchical logistic regression example, showing the expressiveness of the modelling approach as well as the ease of running the parameter inference. We also perform an empirical evaluation of the runtime of the private inference on a complex model and find a $\sim$10 fold speed-up compared to an implementation using TensorFlow Privacy.
}\end{abstract}


%%% main text body
%%%%%%%%%%%%%%%%%%%% INTRODUCTION %%%%%%%%%%%%%%%%%%%%%%%%%%%%%%%%%%%%%%
\section{Introduction}

Probabilistic modelling presents a natural way to model data by describing their (assumed) generative process. The model is then fit to observations by probabilistic inference algorithms. Probabilistic programming aims to make the process easy by allowing the user to only specify the model while the system manages the inference process. Probabilistic programming frameworks such as Stan \cite{carpenter2017stan}, Pyro \cite{bingham2018pyro} and PyMC3 \cite{fonnesbeck2015pymc} have become popular, but they currently offer no support for privacy-preserving algorithms, which are needed for learning from sensitive data.

Differential privacy (DP) \cite{dwork2006calibrating} provides a rigorous mathematical framework for addressing privacy concerns and has become the de-facto standard notion for privacy in machine learning. It essentially assures that an algorithm's outputs will not differ significantly whether a specific individual's data record is included in the data set or not. Unfortunately, differentially-private algorithms are usually more complex than their non-private counterparts. Software support for easily performing fast differentially-private inference is therefore a crucial tool to achieve privacy-preserving probabilistic programming. This will greatly simplify applications such as differentially-private data anonymisation using a generative probabilistic model to publish a privacy-preserving synthetic twin of a sensitive data set \cite{jalko2019privacy}.

Using existing probabilistic programming frameworks with privacy-preserving inference is a highly non-trivial task.
Practitioners are forced to come up with their own implementation, either from scratch or by adapting existing privacy-enabling libraries, which can be an onerous process and leads to many users having to implement the same (or quite similar) wrapper code. There is therefore a clear need for software solutions that enable privacy-preserving probabilistic programming in a convenient and runtime efficient way to allow for fast prototyping and development involving probabilistic programming under privacy constraints.

We address this gap and extend the tool set for growing adoption of DP by introducing an open-source Python software package called \dppp.\footnote{Available at: \url{https://github.com/DPBayes/d3p}} \dppp focuses on providing a reliable high-performance implementation of differentially-private doubly stochastic variational inference (DP-VI) \cite{jalko2016differentially} for tabular data, where each record corresponds to a single individual. \dppp extends the NumPyro probabilistic programming framework \cite{phan2019composable, bingham2018pyro}, allowing modellers to express and fit a large class of probabilistic models under strict privacy guarantees. Alongside the fundamental DP-VI algorithm, \dppp uses a state-of-the-art privacy accounting technique \cite{koskela2020computing} to compute tight bounds on the privacy parameters, allowing it to achieve higher levels of utility than with other commonly employed accountants. 

Behind the scenes, \dppp relies on the JAX framework \cite{bradbury2018jax} to perform computation on GPUs and implements a GPU-optimised minibatch sampling algorithm to further optimise performance.
Using \dppp we achieve a $\sim$10 fold speedup for fitting a variational auto-encoder model \cite{kingma2013auto} compared to a similar implementation using TensorFlow Privacy \cite{tensorflow2019privacy} on modern GPUs.

\dppp addresses a research audience of probabilistic modelling practitioners working with sensitive data. We aim to provide a helpful tool for experimental modelling under privacy constraints. Our main focus in its design therefore is on usability and runtime performance to enable fast modelling iterations. Due to this, \dppp currently does not address technical issues arising from implementing idealised differentially private algorithms on machines with imperfect sources of randomness and finite precision, discussed further in Sec.~\ref{sec:PracticalDPBackground}; these could theoretically be exploited by an adversary if deployed in a production setting.

In summary, we contribute a versatile and performant off-the-shelf implementation of a privacy-preserving probabilistic programming framework as a solid basis for further research. Additionally, we introduce a highly performant subsampling approach based on a slight modification of the CUDA-Shuffle \cite{mitchell2021gpushuffling}, a recently introduced GPU-optimised shuffling approach, and provide a (probabilistic) runtime analysis for it as a minor contribution in methods.

The remainder of the paper is organised as follows: Section~\ref{sec:background} reviews probabilistic programming and differentially private variational inference. Based on that discussion we identify software requirements for our software package to clearly outline our design considerations in the same section. We then demonstrate use of our software on a non-trivial hierarchical logistic regression example, illustrating the expressiveness of the probabilistic programming approach and probabilistic modelling (Section~\ref{sec:usageExample}). Section~\ref{sec:implementation} highlights some implementation details that are orthogonal to the private inference algorithm that forms the core of our framework but that we consider interesting for the user. This includes the discussion of and establishing of (probabilistic) runtime bounds for a special case of the CUDA-Shuffle algorithm which enables GPU-optimised minibatch sampling in our software.  Finally, Section~\ref{sec:results} presents an evaluation of the \dppp framework, including a runtime comparison to a TensorFlow-based implementation, a demonstration of the model introduced in Section~\ref{sec:usageExample} and a replication of an experiment for the DP-VI algorithm in \cite{jalko2016differentially}.

%%%%%%%%%%%%%%%%%%%% BACKGROUND %%%%%%%%%%%%%%%%%%%%%%%%%%%%%%%%%%%%%%

\section{Differentially Private Probabilistic Programming}
\label{sec:background}

In this section, we review the background and techniques for differentially private probabilistic programming that inform the implementation choices of our framework. We start with a broad general introduction of the probabilistic programming paradigm and variational inference (Sec.~\ref{sec:PPBackground}) and the definition of differential privacy as our main privacy formalism (Sec.~\ref{sec:DPBackground}). Following that we give an outline of the powerful DP-VI private inference algorithm (Sec.~\ref{sec:DPSVIBackground}) and review privacy accounting tools (Sec.~\ref{sec:accountantsBackground}). Finally we briefly point out technical difficulties resulting from implementing idealised differentially private algorithms on machines with imperfect sources of randomness and finite precision (Sec.~\ref{sec:PracticalDPBackground}). Each subsection provides an overview of the topic and allows us to identify major requirements for a software implementation, which are summarised in Table~\ref{tab:requirements}. The requirements we identify correspond directly to our overarching goals of providing software that is convenient to use and highly performant. Accordingly, we categorise requirements in Table~\ref{tab:requirements} with the labels \emph{usability} or \emph{performance}. We hope that explicitly stating our design goals here will allow the reader to evaluate whether our design goals are suitable for their use case and make an informed decision on whether to use \dppp. We also do so to emphasise that the implementation of a software package for general use must consider other factors than a (prototypical) implementation of a newly devised method for a research paper.

\begin{table*}[htb!]
    \centering
    \begin{tabular}{c|l|c|c}
        \toprule
         \bfseries No. & Requirement & Category & Section \\
         \midrule
         \reqentry{req:integrate} & Integrate with an existing popular probabilistic programming framework. & \emph{usability} & \ref{sec:PPBackground} \\
         \reqentry{req:privacyBoundsGuidance} & Provide assistance for the user in finding adequate privacy bounds & \emph{usability} & \ref{sec:DPBackground} \\
         \reqentry{req:instanceGrads} & Perform efficient per-instance gradient computation and clipping. & \emph{performance} & \ref{sec:DPSVIBackground} \\
         \reqentry{req:dpsviParams} & Determine DP inference algorithm parameters ($C$ and $\sigma$) automatically. & \emph{usability} & \ref{sec:DPSVIBackground} \\
         \reqentry{req:batchSampling} & Perform efficient independent minibatch sampling. & \emph{performance} & \ref{sec:DPSVIBackground} \\
         \reqentry{req:privacyAccountant} & Provide state-of-the-art privacy accounting. & \emph{usability} & \ref{sec:accountantsBackground} \\
         \bottomrule
    \end{tabular}
    \caption{Requirements for the differentially private probabilistic programming framework with their corresponding category and the subsection they were discussed in (in order of appearance in the text).}
    \label{tab:requirements}
\end{table*}

\subsection{Probabilistic Programming}
\label{sec:PPBackground}

Probabilistic programming is a programming paradigm in which a user programmatically defines a statistical model of data which often depends on a set of parameters $\theta$. In mathematical terms, such a model determines a probabilistic density function $p(\cdot|\theta)$. A probabilistic inference algorithm is then used to determine the posterior distribution $p(\theta|X)$ of the parameter values given a training data set $X$. The posterior is given by Bayes' formula as $p(\theta|X) \propto p(X|\theta) p(\theta)$, where, $p(X|\theta)$ is the likelihood of the data under the probabilistic model with given parameter values. $p(\theta)$ is a prior distribution encapsulating existing knowledge about plausible parameter values.

There exist a number of different probabilistic programming languages that use different ways of specifying the model. To enable easy user adoption, a software package providing differential privacy for probabilistic programming should not aim to re-define and re-implement yet another solution but rather integrate with existing solutions by extending them with support for differentially private inference (Table~\ref{tab:requirements}, Requirement~\ref{req:integrate}).

\paragraph*{Defining an example model}

To illustrate the concept of probabilistic programming, we present as an example the implementation of logistic regression for binary classification in NumPyro.

The simple logistic regression model we consider first is for a data set $X$ of records $\bm{x}_i \in \mathbb{R}^D$ with corresponding labels $y_i \in \{0, 1\}$. We assume that each such record and label corresponds to a single individual.

Mathematically the logistic regression model is formulated as
\begin{align}
    p(y_i|\bm{x}_i, \bm{w}) &= \text{Bernoulli}\left(y_i; \theta_i\right), \nonumber \\
    \theta_i &= \sigma(\bm{w}^T \bm{x}_i), \nonumber \\
    p(\bm{w}) &= \mathcal{N}(\bm{w}; \bm{w}_0, \bm{S}_0), \nonumber
\end{align}
where $\sigma(x) = 1/(1+\exp(-x))$ denotes the sigmoid function and $\text{Bernoulli}(\cdot; \theta)$ denotes the Bernoulli distribution with success probability $\theta$. The model uses a weight vector $\bm{w} \in \mathbb{R}^D$ to express the relationship between the records and labels. Bayesian treatment allows us to formulate a prior on $\bm{w}$ to express any prior knowledge about plausible parameter values. Here we use a weakly informative, zero-centered Gaussian prior with $\bm{w}_0 = 0$ and $\bm{S}_0 = 4\bm{I}$. The mathematical description of the model equations translates naturally into Python code for NumPyro model definition in Listing \ref{lst:model}.

\begin{lstlisting}[
    float=h, language=Python, label=lst:model,
    caption={Definition of a simple logistic regression model in NumPyro for \dppp.}
]
# specifies the model p(ys, w | xs)
def model(xs, ys, N): 
    # obtain data dimensions
    batch_size, d = xs.shape
    
    # the prior for w
    w = sample('w', Normal(0, 4),
               sample_shape=(d,))
    
    # distribution of label y for each record x
    with plate('batch', N, batch_size):
        theta = sigmoid(xs.dot(w))
        sample('ys', Bernoulli(theta), obs=ys)
\end{lstlisting}

\paragraph*{Doubly Stochastic Variational Inference}

At the heart of probabilistic programming lies the inference algorithm that is used to determine the posterior distribution of parameters $p(\theta|X)$. For complex models computing this posterior exactly is typically intractable. Variational inference \cite{jordan1999introduction, wainwright2008graphical}, a class of approximate inference algorithms, therefore approximates it with a simpler, tractable distribution $q(\theta|\psi)$. The parameters $\psi$ of $q$ are found by solving the optimisation problem $\min_{\psi} D(q(\theta|\psi) || p(\theta|X))$ where $D(\cdot||\cdot)$ is a divergence measure for probability distributions.

Doubly stochastic variational inference (DSVI) \cite{titsias2014doubly} is a gradient ascent algorithm for non-conjugate models with differentiable (joint) probability densities $p(X,\theta)$ and a $q(\theta|\psi)$ from which values can be easily sampled algorithmically. While these conditions limit applicability somewhat, they still allow for a large class of models to be fitted.

DSVI minimises the KL-divergence by maximising the so-called \emph{evidence lower bound (ELBO)}, defined as
\begin{equation}
    \mathcal{L}(\psi|X) = \mathbb{E}_{\theta \sim q(\theta|\psi)} \left[\log p(X|\theta) + \log p(\theta) - \log q(\theta|\psi)\right].
\end{equation}

The expectation is approximated stochastically by sampling $\theta$ from $q(\theta|\psi)$ and using a minibatch of the training data for each gradient step. For details we refer to \cite{titsias2014doubly}.

\subsection{Differential Privacy}
\label{sec:DPBackground}

We rely on (approximate) differential privacy \cite{dwork2006calibrating, dwork2006our} as the primary privacy notion for a privacy-preserving variant of the DSVI algorithm. Following \cite[Def.~2.4]{dwork2014algorithmic}, it is defined as:

\begin{definition}[Approximate Differential Privacy]
\label{def:dp}
A randomised algorithm $\mathcal{M}$ satisfies $(\varepsilon, \delta)$-differential privacy with $\varepsilon >0$ and $0 \leq \delta \leq 1$ if, for all neighbouring data sets $X \sim X'$, and for all $S \subset \text{im}(\mathcal{M})$, we have
\begin{align}
    \Pr(\mathcal{M}(X) \in S) \leq e^\varepsilon\Pr(\mathcal{M}(X') \in S) + \delta.
\end{align}
\end{definition}

The two data sets $X$ and $X'$ are considered to be neighbours, denoted $X \sim X'$, when we can obtain one from the other by adding (resp.~removing) a single element. $\varepsilon$ and $\delta$ are \emph{privacy bounds} (or \emph{privacy parameters}) restricting the effect that the presence of any particular record in the input data set has on the output of the algorithm $\mathcal{M}$.

Smaller values for these privacy bounds correspond to stricter privacy, however there is a trade-off between privacy and utility of the algorithm's outputs. Larger values for privacy bounds typically result in higher utility of the outputs, as they allow more information to pass through the algorithm. Choosing the privacy bounds therefore requires careful consideration of this trade-off. This is difficult because for many users, especially those inexperienced with DP, it is not clear how to interpret the privacy bounds in a concrete setting. The software should therefore assist the user in choosing appropriate privacy bounds which we reflect in Requirement \ref{req:privacyBoundsGuidance}.

\subsection{Differentially Private Doubly Stochastic Variational Inference}
\label{sec:DPSVIBackground}

\citeauthor{jalko2016differentially} introduced a $(\varepsilon, \delta)$-DP version of the doubly stochastic variational inference algorithm in \cite{jalko2016differentially}. This DP-VI algorithm is derived from the influential DP-SGD \cite{song2013stochastic, abadi2016dpsgd} and the relevant steps of a single iteration (out of $T$ many) can be summarised as

\begin{enumerate}
    \item Sample a random minibatch of size $B$ from the training data set.
    \item Sample a set of parameters $\theta$ from $q(\cdot|\psi)$.
    \item For each instance in the minibatch: \begin{enumerate}[label=\arabic*.]
        \item Compute the gradient of the ELBO.
        \item Clip the norm of the gradient to a bound $C$.
    \end{enumerate}
    \item Aggregate per-instance gradients.
    \item Perturb by adding zero-mean Gaussian noise with variance $C^2 \sigma^2$.
    \item Update the model parameters $\psi$ with the perturbed gradient.
\end{enumerate}

The main mechanism by which differential privacy is achieved is the perturbation of the minibatch gradient in step 5 via the Gaussian mechanism \cite[Thm.~3.22]{dwork2014algorithmic}. The level of noise, characterised by its variance $\sigma^2$, must be carefully calibrated to provide the desired level of privacy. However, the gradient of any data instance could in theory be arbitrarily large, rendering any fixed noise level ineffective. To remedy this, the DP-VI algorithm enforces an upper bound $C$ on the gradient of each data instance (in step 2.2). The important implication of this is that an implementation of DP-VI needs an efficient way of computing and manipulating the per-instance (often also known as per-example) gradients in a minibatch instead of a single gradient over the entire minibatch (Requirement~\ref{req:instanceGrads}).

Another important observation is that the DP-VI algorithm has additional hyperparameters $C$ and $\sigma$ which govern the privacy vs.~accuracy trade-off. Especially $\sigma$ depends non-trivially on the clipping bound $C$, desired privacy bounds $\varepsilon$ and $\delta$, batch size $B$ and the number of iterations $T$. The next requirement for the software package is therefore the ability to (automatically) derive appropriate values for the DP-VI hyperparameters from these other hyperparameters (Requirement~\ref{req:dpsviParams}).

A final crucial point is that the algorithm is shown to provide differential privacy only under the assumption that minibatches are independently sampled from the training set. As this needs to occur in every iteration of the algorithm, this routine must be especially fast to not slow down the inference as a whole, making another performance requirement for our software (Requirement~\ref{req:batchSampling}).

\subsection{Privacy Accounting}
\label{sec:accountantsBackground}

As we have seen, the DP-VI algorithm consists of an iterative application of the (subsampled) Gaussian mechanism on gradients wrt.~random minibatches of the training data. The overall privacy bounds $\varepsilon$ and $\delta$ of the DP-VI algorithm then result from the DP composition. In order to achieve good utility it is crucial to compute these overall privacy bounds to be as tight as possible: Looser bounds mean that larger perturbations are required for a desired level of privacy, reducing the information extracted from the data and decreasing utility of the inferred model (cf.~\cite{erlingsson2019that}).

While loose bounds can be computed using general DP composition theorems in a simple way (cf.~\cite{dwork2014algorithmic}), obtaining tight bounds typically requires more complex computation using methods called \emph{privacy accountants}. \citeauthor{abadi2016dpsgd}'s Moments accountant \cite{abadi2016dpsgd} was the first of these and significantly improved over traditional DP composition theorems. The tightest privacy bounds are currently achieved by the Fourier accountant \cite{koskela2020computing}. 

Privacy accountants are typically of the form $f_{PA}(C, \sigma, B, T, \delta) = \varepsilon$, i.e., they take in the algorithm's parameters as well as a target value for $\delta$ and compute the corresponding upper bound for $\varepsilon$. They are therefore the primary tool to translate between privacy bounds and inference hyperparameters and instrumental for addressing Requirements \ref{req:privacyBoundsGuidance} and \ref{req:dpsviParams}.

With these considerations, providing an implementation of a state-of-the-art privacy accountant is an important aspect for a differentially private probabilistic programming framework and becomes Requirement~\ref{req:privacyAccountant}. 

\subsection{Remaining Technical Concerns for the Practical Implementation}
\label{sec:PracticalDPBackground}

The definition of approximate differential privacy given in Section~\ref{sec:DPBackground} provides information-theoretic guarantees: It is impossible for the output probabilities of an algorithm to vary too much no matter what the input is. These can typically not be achieved by a computer system which do not have access to perfect sources of randomness for sampling noise and rely on finite-precision approximation of real numbers. Both of these have the potential to completely void the privacy guarantees of DP algorithms in practical implementations: Predictable randomness can allow an attacker to remove the perturbations \cite{garfinkel2020randomness} and finite-precision floating point numbers can leak information due to approximation errors \cite{mironov2009computational}.

We recognise these issues as generally important for production systems. However, as we have already pointed out, we undergo no effort to address these for the current version of \dppp, which is primarily intended as a research tool. We consider solving these issues for the DP-VI algorithm as important future work.

\subsection{Summary}
We have seen in this section that there is a large number of desiderata for an implementation of a differentially private probabilistic programming framework. These are summarised in Table~\ref{tab:requirements} and fall into different categories that make some mostly technical considerations for the implementation (e.g., Requirements~\ref{req:instanceGrads}, \ref{req:batchSampling}), while others are important aspects in the design of the user interface (e.g., Requirements~\ref{req:privacyBoundsGuidance}, \ref{req:dpsviParams}).

In the following sections we first explore how our software package addresses these requirements from a user perspective by implementing an example model. However, some of the technical requirements are not experienced by the user in the programming interface directly and cannot be demonstrated in the example: Whether the implementation is performant (Requirements~\ref{req:instanceGrads}, \ref{req:batchSampling}) has almost no effect on the interface seen by the user but is nevertheless a crucial part of their experience. We therefore briefly discuss some implementation details following the examples to illustrate how the identified requirements were addressed and, following that, provide an empirical evaluation of the runtime performance.

%%%%%%%%%%%%%%%%%%%% EXAMPLE %%%%%%%%%%%%%%%%%%%%%%%%%%%%%%%%%%%%%%

\section{\dppp Usage Example}
\label{sec:usageExample}

We will now demonstrate our \dppp software package on a practical example to show how the previously identified requirements are addressed from a user's perspective. \dppp centers around  an efficient implementation of the DP-VI algorithm and therefore admits non-conjugate models with differentiable probability densities. It is designed to provide differential privacy guarantees for tabular data, where each individual contributed a single sensitive record and records are assumed conditionally independent. \dppp uses NumPyro \cite{phan2019composable, bingham2018pyro} as a modelling language and JAX \cite{bradbury2018jax, frostig2018compiling} as the underlying computation framework, which offers an API similar to NumPy \cite{harris2020array}.
We start by fully implementing the simple logistic regression model shown in Section~\ref{sec:PPBackground} to demonstrate the basics of probabilistic programming and show how \dppp's DP-VI algorithm is invoked to infer the model's parameters. We then highlight the expressiveness of the probabilistic programming approach by adapting the code to a more complex model that achieves a better fit to the data.

\subsection{Defining a Model}

We recall the simple logistic regression model introduced in Sec. \ref{sec:PPBackground}. The model is for a data set $X$ of records $\bm{x}_i \in \mathbb{R}^D$ with corresponding binary labels $y_i \in \{0, 1\}$. We assume that each such record and label corresponds to a single individual.

\begin{lstlisting}[
    float=tb, language=Python, label=lst:fullmodel,
    caption={Implementation of a simple logistic regression model in NumPyro for \dppp.}
]
import jax.numpy as jnp

# specifies the model p(ys, w | xs)
def model(xs, ys, N): 
    # obtain data dimensions
    batch_size, d = xs.shape 
    
    # the prior for w
    w = sample('w', Normal(0, 4),
               sample_shape=(d,))
    
    # distribution of label y for each record x
    with plate('batch', N, batch_size):
        theta = sigmoid(xs.dot(w))
        sample('ys', Bernoulli(theta), obs=ys)
        
# specifies the variational posterior q(w)
def guide(xs, ys, N):
    d = jnp.shape(xs)[1]
    
    # variational parameters
    w_loc = param('w_loc', jnp.zeros((d,))
    w_scale = jnp.exp(param('w_scale_log',
                            jnp.zeros((d,)))
    
    # variational distribution for w
    sample('w', Normal(w_loc, w_scale))
\end{lstlisting}

\begin{figure}[h]
    \begin{center}
    \begin{tikzpicture}
      % nodes
        \node[obs] (y) {$y_{i}$};%
        
        \node[obs, below=of y] (x) {$\bm{x}_{i}$};%
        \plate [inner sep=.3cm] {observations} {(x)(y)} {$N$};%
        
        \node[latent, left=of y] (w) {$\bm{w}$};%
        \node[det, left=of w] (w0) {$\bm{w}_0$};%
        \node[det, below=of w0] (S0) {$\bm{S}_0$};%
      % edges
        \edge {x, w} {y};%
        \edge {w0, S0} {w};
    \end{tikzpicture}
    \end{center}
    \caption{Plate diagram of the simple logistic regression model $p(y, \bm{w}|\bm{x})$.}
    \label{fig:simpleModelPlate}
\end{figure}
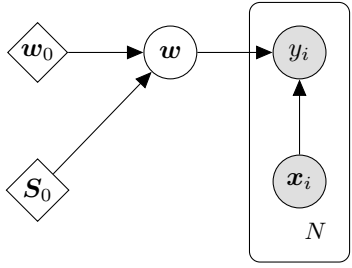

As before, the model can be formulated mathematically as
\begin{align}
    p(y_i|\bm{x}_i, \bm{w}) &= \text{Bernoulli}\left(y_i; \theta_i\right), \label{eq:logreg_y_repeat} \\
    \theta_i &= \sigma(\bm{w}^T \bm{x}_i), \nonumber \\
    p(\bm{w}) &= \mathcal{N}(\bm{w}; \bm{w}_0, \bm{S}_0), \nonumber
\end{align}
where $\sigma(x) = 1/(1+\exp(-x))$ denotes the sigmoid function and $\text{Bernoulli}(\cdot; \theta)$ denotes the Bernoulli distribution with success probability $\theta$. The weight vector $\bm{w} \in \mathbb{R}^D$ links the records and labels. For a Bayesian treatment we place a prior on $\bm{w}$ to encode existing knowledge. In this first example we assume that we do not have strong prior knowledge but want to enforce some regularisation, and therefore use a weakly informative, zero-centered Gaussian prior with $\bm{w}_0 = 0$ and $\bm{S}_0 = 4\bm{I}$. The model is visualised using plate diagram notation in Figure~\ref{fig:simpleModelPlate}.

\paragraph*{Implementation of the Model}
The implementation of this model using the NumPyro probabilistic programming framework is reproduced in the \texttt{model} function in the top part of Listing~\ref{lst:fullmodel}. The model is defined as a function taking (a minibatch of) the data as an input and specifies a sampling process in an imperative programming style: Values for $\bm{w}$ are sampled from the specified Gaussian prior with zero mean and a standard deviation of $4$. The values for labels are sampled from the Bernoulli distribution (cf.~Eq.~\ref{eq:logreg_y_repeat}). However, the \texttt{sample} call for $y$ is conditioned to return the values \texttt{ys} passed into the \texttt{model} function using the \texttt{obs} keyword. This is the mechanism by which the labels are passed into the inference algorithm despite the model being specified from a generative perspective.

NumPyro's \texttt{plate} context manager is used to express the independence assumption for the individual data records. Note that this is an important assumption in the \dppp package and must be reflected in the model in this way.\footnote{Apart from clearly stating the assumptions made for the data, this ensures that using minibatches instead of the whole data set does not affect the amount by which an individual sample contributes to the ELBO.} This requires the additional argument \texttt{N} to the model, which specifies the total amount of data records in the training data.

\begin{lstlisting}[
    float=tb, language=Python, label=lst:inference,
    caption={Running the inference for a NumPyro model using \dppp's \texttt{DPSVI} class.}
]
def infer(data, labels, batch_size, num_iter, epsilon, delta, rng_key):
    # set up minibatch sampling
    batchifier_init, get_batch = \
        subsample_batchify_data((data, labels),
                                batch_size)
    _, batchifier_state = \ 
        batchifier_init(rng_key)
    
    # set up DP-VI algorithm
    q = batch_size / len(data)
    dp_scale, _, _ = approximate_sigma(
        epsilon, delta, q, num_iter)
    loss = Trace_ELBO()
    optimiser = Adam(1e-3)
    dpsvi = DPSVI(model, guide, optimiser,
                  loss, dp_scale, len(data))
    svi_state = dpsvi.init(
        rng_key,
        *get_batch(0, batchifier_state))
    
    # run inference
    for i in range(num_iter):
        data_batch, label_batch = \ 
            get_batch(i, batchifier_state)
        svi_state, loss = dpsvi.update(
            svi_state, data_batch, label_batch)
    return dpsvi.get_params(svi_state)
\end{lstlisting}

\paragraph*{Implementation of the Variational Posterior}
For inference of the model's parameters in our example, we use independent Gaussian distributions for every data dimension $j$ with parameters $\mu_{\bm{w},j}$ and ${\sigma}_{\bm{w}, j}$ as the variational approximation to the posterior distribution, i.e.,
\begin{equation}
    q({w}_j | \mu_{\bm{w}, j}, \sigma_{\bm{w}, j}) = \mathcal{N}({\mu}_{\bm{w},j}, {\sigma}_{\bm{w},j}^2). \label{eq:guide}
\end{equation}
The variational parameters $\bm{\mu}_{\bm{w}}$ and $\bm{\sigma}_{\bm{w}}$ will be optimised according to the discussion in Section~\ref{sec:PPBackground}. The corresponding NumPyro implementation is shown in the \texttt{guide} function, following naming conventions of NumPyro, in the lower portion of Listing~\ref{lst:model}. We register $\bm{\mu}_{\bm{w}}$ and $\bm{\sigma}_{\bm{w}}$ as parameters \texttt{w\_loc} and \texttt{w\_scale} for the inference algorithm and sample $w$ according to Equation~\ref{eq:guide} in a vectorised fashion. Note that each $\sigma_{\bm{w}, j}$, the standard deviation of the variational Gaussian, must be a positive number, so we actually register a parameter site named \texttt{w\_scale\_log} that we pass through the exponential function to obtain \texttt{w\_scale}. This allows us to perform the optimisation in an unconstrained space but enforces the positivity constraint for \texttt{w\_scale}.

\subsection{Running the Inference}

In the previous section we programmatically specified the simple logistic regression model using NumPyro. We now turn to the actual private inference of parameter values using \dppp. \dppp provides an implementation of the DP-VI algorithm via the \texttt{DPSVI} class, which offers the same interface as NumPyro's implementation of the DSVI algorithm in the \texttt{SVI} class.\footnote{We chose \texttt{DPSVI} as the name for implementation of the DP-VI algorithm in \dppp to stay close to NumPyro's naming convention.} As discussed in Section~\ref{sec:DPSVIBackground}, the DP-VI algorithm must be configured for the desired privacy bounds given the batch size and number of training iterations for the inference (Req.~\ref{req:dpsviParams}) and use independent random minibatches (Req.~\ref{req:batchSampling}). The entire code for running the inference is shown in Listing~\ref{lst:inference}.

We first use \dppp's \texttt{subsample\_batchify\_data} on the data set which returns a function that efficiently samples and returns independent random minibatches. This function is assigned to \texttt{get\_batch} in our example code. The additional call to \texttt{batchify\_init} initialises the internal state of the minibatch sampler.

To instantiate a \texttt{DPSVI} object, the main driver of the inference, we need to supply a value for the privacy noise scale $\sigma$. We can obtain a $\sigma$ appropriate for our desired privacy bounds and training hyperparameters using the \texttt{approximate\_sigma} function. This function returns an approximate value for $\sigma$ that is guaranteed to achieve the privacy bounds as measured by the Fourier accountant \cite{koskela2020computing}, a state-of-the-art privacy accountant method. We store the result in \texttt{dp\_scale}. Note that the separate computation of $\sigma$ is a deliberate choice in \dppp. While it would be possible to let the instantiation code of the \texttt{DPSVI} class handle this internally, the current approach allows the user easily provide values for $\sigma$ different from the ones computed by the \texttt{approximate\_sigma} function, e.g., for research purposes.

After also instantiating implementations of the ELBO (\texttt{loss}) and an optimiser of our choice (\texttt{optimiser}) using classes provided by NumPyro, we are ready to create the \texttt{DPSVI} object (\texttt{dpsvi}). Similar to the minibatch sampler, the DP-VI algorithm provides an initialisation function that is called to produce a state object (\texttt{svi\_state}). The state object contains randomness state as well as the current values of the parameters and the state of the optimiser. We can now finally run the inference by repeatedly sampling a batch using the \texttt{get\_batch} function we obtained before and then calling \texttt{dpsvi.update}. The \texttt{update} method completely encapsulates a single iteration of the DP-VI algorithm, including the performant per-instance gradient computation (Req.~\ref{req:instanceGrads}), clipping, perturbation and the update of current parameter estimates by the optimiser.

Obtaining parameter estimates from the inference algorithm (and therefore the approximate posterior distribution $q(w)$ in our example model) completes our example at this point. The user can now use standard NumPyro code to interact with the model and the inferred parameters without additional privacy leakage due to DP's invariance to post-processing.

\subsection{Switching to a More Complex Model}
\label{sec:ComplexExample}

One of the main benefits of the probabilistic programming approach is the ability to easily tailor the model complexity to the information needs and the available prior knowledge and clearly specifying how the respective components of the model relate to each other. So far we have looked at a very simple example where our model makes the implicit assumption that the data are homogeneous and a single parameter vector $\bm{w}$ describes their relation to the labels equally well for all records. In reality, however, we often face tasks where data comes from different sources that have different local distributions for records. For example, data records containing information about wealth and income of individual persons from different countries are likely to be heavily influenced by the average level of wealth in the respective country. To address this case, we can use a hierarchical logistic regression model like the one considered in \cite{wong1985hierarchical}.

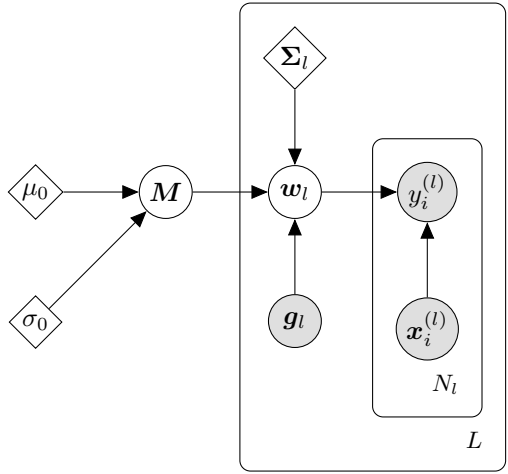
\begin{figure}[h]
    \begin{center}
    \begin{tikzpicture}
      % nodes
        \node[obs] (y) {$y_{i}^{(l)}$};%
        
        \node[obs, below=of y] (x) {$\bm{x}_{i}^{(l)}$};%
        \plate [inner sep=.3cm] {observations} {(x)(y)} {$N_l$};%
        
        \node[latent, left=of y] (w) {$\bm{w}_l$};%
        \node[obs, below=of w] (g) {$\bm{g}_l$};%
        \node[det, above=of w] (Sl) {$\bm{\Sigma}_l$};%
        \plate [inner sep=.3cm] {groups} {(observations)(w)(Sl)} {$L$};%
        \node[latent, left=of w] (M) {$\bm{M}$};%
        
        \node[det, left=of M] (mu0) {$\mu_0$};%
        \node[det, below=of mu0] (sig0) {$\sigma_0$};%
        
      % edges
        \edge {x, w} {y};%
        \edge {g, M, Sl} {w};
        \edge {mu0, sig0} {M};
    \end{tikzpicture}
    \end{center}
    \caption{Plate diagram of the hierarchical logistic regression model $p(y, \bm{M} |\bm{x}, \bm{g}_l)$.}
    \label{fig:hierarchicalModelPlate}
\end{figure}

\paragraph*{Extending the Model Specification}
We now extend the previous model notation by vectors $\bm{g}_l \in \mathbb{R}^K$ of group characteristics for $L$ groups and, for each record $\bm{x}_i$, an indicator $l_i$ assigning it to one of the groups. We assume that the records $\bm{x}_i$ and labels $y_i$ are sensitive but the group vectors $\bm{g}_l$ are not. Following \cite{wong1985hierarchical} we use a separate weight vector $\bm{w}_l$ to model the relationship between data record $\bm{x}_i$ and label $y_i$ within each group in the same way we did in the simple logistic regression model. However, we now use a hierarchical Gaussian prior centered at $\bm{M}\bm{g}_l$ for each weight vector $\bm{w}_l$. The matrix $\bm{M} \in \mathbb{R}^{D \times K}$ is a new parameter capturing the relation between group characteristics and their corresponding weight vector $\bm{w}_l$. The full model is visualised in Figure \ref{fig:hierarchicalModelPlate} and defined by

\begin{align}
    p(y_i^{(l)}|\bm{x}_i^{(l)}, \bm{w}_l) &= \text{Bernoulli}(y_i^{(l)}; \theta_i^{(l)}), \label{eq:hiera_logreg_y} \\
    \theta_i^{(l)} &= \sigma(\bm{w}_l^T \bm{x}_i^{(l)}), \nonumber \\
    p(\bm{w}_l | \bm{g}_l, \bm{M}, \bm{\Sigma}_l) &= \mathcal{N}(\bm{w}_l; \bm{\eta}_l, \bm{\Sigma}_l), \nonumber \\
    \bm{\eta}_l &= \bm{M} \bm{g}_l, \nonumber \\
    p(M_{k,d}) &= \mathcal{N}(M_{k,d}; \mu_0, \sigma_0^2). \nonumber
\end{align}

For simplicity, we consider the covariance matrix $\bm{\Sigma}_l$ for the distribution of the $\bm{w}_l$ a fixed model parameter of value $\bm{\Sigma}_l = \bm{I}$. We assume an independent weakly informative Gaussian prior for each element of matrix $\bm{M}$ with $\mu_0 = 0$ and $\sigma_0 = 4$ as before.

\begin{lstlisting}[
    float=tb, language=Python, label=lst:complexModel,
    caption={Implementation of a hierarchical logistic regression model in NumPyro for \dppp.}
]
def model(xs, ys, ls, gs, N):
    batch_size, D = xs.shape
    L, K = gs.shape
    
    M = sample('M', Normal(0, 4),
               sample_shape=(D, K))
    
    with plate('group', L, L):
        etas = gs @ M.T
        ws = sample(
            'ws', Normal(etas, 1).to_event(1))
        
    with plate('batch', N, batch_size):
        thetas = sigmoid(jnp.einsum(
            "nd,nd->n", xs, ws[ls]))
        sample('ys', Bernoulli(thetas), obs=ys)
        
def guide(xs, ys, ls, gs, N):
    _, D = xs.shape
    _, K = gs.shape
    
    M_loc = param('M_loc', jnp.zeros((D, K)))
    M_scale = jnp.exp(param('M_scale_log',
                            jnp.zeros((D, K))))
    sample('M', Normal(M_loc, M_scale))
\end{lstlisting}

\paragraph*{Model Implementation}
The required changes to adapt our existing model implementation in function \texttt{guide} are straightforward and essentially follow one to one from the textual description above, as shown in Listing~\ref{lst:complexModel}. First we sample a value for $\bm{M}$ according to our prior (\texttt{M}), compute the values for $\bm{\eta}_l$ (\texttt{etas}) and sample the group-specific weight vectors $\bm{w}_l$ (\texttt{ws}) in a vectorised manner. The remaining code implementing the logistic regression for individual records is almost identical to the implementation of the simple model, except that we use the group indicators $l_i$ provided in \texttt{ls} to select the entry in \texttt{ws} that corresponds to the $\bm{w}_{l_i}$ of the group each record belongs to.

\paragraph*{Variational Posterior}
For the hierarchical logistic regression model, we are interested in a variational approximation to the posterior of the matrix $\bm{M}$. Again we can use independent Gaussian distributions for each dimension of $\bm{M}$:
\begin{equation}
    q(M_{k,d} | \mu_{kd}, \sigma_{kd}) = \mathcal{N}(M_{k,d} ; \mu_{kd}, \sigma_{kd}^2),
\end{equation}
where $\mu_{kd}$ and $\sigma_{kd}$ for all $1 \leq k \leq K$ and $1 \leq d \leq D$ are the variational parameters.

The implementation of this is shown in the \texttt{guide} function in Listing~\ref{lst:complexModel} and is almost identical to the one for the simple model. Invocation of the inference algorithm for the new model also does not change compared to the simple logistic regression model we have looked at before, except for passing the additional data to the \texttt{model} and \texttt{guide} functions, making it especially easy and fast to refine and tweak models.

Note that we do not specify posterior parameters for the group weight vectors $\bm{w}_l$. This is because in our model these are determined by $\bm{M}$, the group characteristics $\bm{g}_l$ and the known covariance matrix $\bm{\Sigma}_l$. If we are interested in obtaining values for the $\bm{w}_l$, we can use NumPyro routines to sample them from the posterior predictive distribution
\begin{equation}
    q(\bm{w}_l | \bm{g}_l) = \int p(\bm{w}_l | \bm{M}\bm{g}_l, \bm{\Sigma}_l) q(\bm{M} | \bm{\mu}, \bm{\sigma}) d\bm{M}
\end{equation}
after inference of the variational parameters for $\bm{M}$. In the above we slightly abuse notation to denote by $q(\bm{M} | \bm{\mu}, \bm{\sigma}) = \prod_{k=1}^K \prod_{d=1}^D q(M_{k,d} | \mu_{kd}, \sigma_{kd})$ the variational posterior of the full matrix $\bm{M}$ for all variational parameters.

\paragraph*{Discussion}
As we have seen we were able to expand our model to the additional structure within our data with a few simple changes in the implementation. The changes correspond directly to the textual description of the new hierarchical model, making it easy for the user to translate theory into implementation. While the initial simple logistic regression model is quite a common choice and specialised implementations for this exist in many software packages, extending the model in a similar fashion to what we did in this section is often more complicated, if at all possible, in those.

Note that we still have made a couple of simplifying assumptions here, one of which is the assumption that the covariance $\bm{\Sigma}_l$ of the distribution of group weight vectors is a known constant. This was merely for convenience and not because of restrictions of model expressiveness. We can easily formulate a prior and a variational posterior for $\bm{\Sigma}_l$ to learn it from the data, if this assumption does not hold. Another simplifying modelling assumption is the choice of the prior for $\bm{M}$, however, we consider selecting more informative priors outside the scope of this example. NumPyro offers a wide range of distributions from which the user can choose adequate priors according to their modelling needs. These two brief notes serve to show that the probabilistic programming paradigm allows the user to make fine-grained decisions on the model expressiveness they need by flexibly either excluding details they are not interested in or incorporating detailed prior knowledge in the model.

%%%%%%%%%%%%%%%%%%%% IMPLEMENTATION %%%%%%%%%%%%%%%%%%%%%%%%%%%%%%%%%%%%%%
\section{\dppp Implementation Outline}
\label{sec:implementation}

We now discuss some technical details of our \dppp package and how it meets the requirements identified in Section~\ref{sec:background}. In line with our discussion of these requirements, we hope that this helps the reader decide whether the choices made in implementing \dppp make it suitable for their use. We also focus on the GPU-optimised minibatch sampling algorithm in particular, which we base on a recent GPU-optimised shuffling method but slightly modify to better fit our use.

We organise this section by first briefly motivating the use of NumPyro as the basis for \dppp (Section~\ref{sec:NumPyroChoice}). We then turn to the implementation of the DP-VI algorithm for NumPyro (Section~\ref{sec:DPSVIImplementation}) with a focus on how it realises its high performance for per-instance gradient computation (Req.~\ref{req:instanceGrads}).

As the second major implementation detail we then discuss our GPU-optimised i.i.d.~minibatch sampling routine (Req.~\ref{req:batchSampling}) which has a major impact on the overall performance of the inference and for which we also contribute a runtime analysis (Section~\ref{sec:gpuSampling}). Finally we finish the discussion on implementation by a brief overview of the remaining usability goals in the implementation (Section~\ref{sec:features}).

\subsection{Underlying Framework}
\label{sec:NumPyroChoice}
To meet Requirement~\ref{req:integrate} (\emph{Integrate with an existing popular probabilistic programming framework}), we chose NumPyro as the underlying probabilistic programming framework for \dppp. It is a spin-off of the popular Pyro \cite{bingham2018pyro} framework, providing a very similar API, but relies on Google's JAX \cite{bradbury2018jax, frostig2018compiling} for the underlying computational optimisation functionality. The JAX framework uses tracing mechanisms to compile pure (side-effect free) functions directly from Python code to efficient XLA kernels that can be run on either CPU or GPU. Implementations of the DP-SGD algorithm in JAX where found to be consistently faster than competing implementations \cite{subramani2020enabling}, making it a promising backend for for \dppp.

The choice of NumPyro as basis for \dppp is therefore a compromise between fulfilling Requirement~\ref{req:integrate} and being able to provide a high-performance implementation (Req.~\ref{req:instanceGrads}, \ref{req:batchSampling}). While targeting Pyro instead of NumPyro would arguably have enabled access to a larger established user base, implementing per-instance gradients in the underlying PyTorch would have decreased performance of \dppp.

\subsection{Implementation of DP-VI}
\label{sec:DPSVIImplementation}

As we have seen in the examples, \dppp centers around the \texttt{DPSVI} class implementing the DP-VI algorithm.  \texttt{DPSVI} offers the same interface as NumPyro's non-private implementation in the \texttt{SVI} class and therefore works as a drop-in replacement for it. This allows for especially easy adoption of privacy-preserving methods with minimal required changes to the code base.

We have identified in Section~\ref{sec:DPSVIBackground} that the DP-VI algorithm requires computation of per-instance gradients. Unfortunately, these are typically not readily available in established machine learning frameworks. A naive but inefficient solution is to set the batch size $B$ to one, which gives per-instance gradients at the cost of losing the performance gains resulting from parallel computation on minibatches that are crucial for efficient machine learning applications.

As the JAX framework is based on the manipulation of side-effect free functions, it also provides a range of composable higher order transformations for these. Crucially, the computation of gradients as well as vectorising a function for parallel execution are examples of these higher order transformations. This enables \dppp's implementation in the \texttt{DPSVI}'s class to efficiently parallelise computation of gradients and the subsequent clipping over a minibatch using vectorisation, i.e., single-instruction-multiple-data (SIMD) style computing. Leveraging the massively parallel processing capabilities of modern GPUs in this way allows us to negate most of the additional overhead introduced by the per-instance gradient computation.

This vectorisation approach effectively turns computation on a batch of size $B$ into $B$ parallel computations on batches of size $1$. In order to prevent this from affecting the relative contributions of prior and variational posterior of global parameters, models must use NumPyro's \texttt{plate} environment to scale likelihood contributions from the batch appropriately to the perceived batch size. This typically means that each data record's contribution to the loss is scaled up. The \texttt{DPSVI} class therefore performs some crucial bookkeeping to ensure that the privacy perturbation in each iteration is also scaled appropriately.

As a convenience feature, the \texttt{DPSVI} class offers methods to obtain tight privacy bounds for its current hyperparameter values which are computed using the Fourier accountant \cite{koskela2020computing}.\footnote{We rely on the \texttt{fourier-accountant} package, available at \url{https://pypi.org/project/fourier-accountant/}, for the implementation)}

\subsection{Performant GPU Batch Subsampling}
\label{sec:gpuSampling}

We have seen in Section~\ref{sec:DPSVIBackground} that the DP-VI algorithm's privacy guarantees rely crucially on minibatches sampled from the data set in a truly i.i.d.~fashion (Requirement~\ref{req:batchSampling}). This is an important difference to non-private algorithms that can usually get away with a permute-and-iterate approach to sample minibatches. Often, the data set is permuted once in its entirety and minibatches are then consumed by iterating over the permuted set. Due to being invoked comparatively rarely, the performance of the permutation algorithm does not make a noticeable difference on the overall runtime. For the same reason, it can also run on a different device than the learning algorithm as slow bus transfers are infrequent.

The i.i.d.~requirement for DP-VI requires that the minibatch sampling routine is invoked once for each iteration, making the cost of a slow sampling routine prohibitive (or at least, much more noticeable). \dppp therefore ships with a parallel and GPU-optimised minibatch sampling routine based on a novel shuffling methodology (CUDA-Shuffle) proposed in \cite{mitchell2021gpushuffling, stokesfeistel}.

Conventional shuffling algorithms, such as the Fisher-Yates shuffle, are ill-suited for GPU-acceleration as they are sequential. In contrast, the main idea in the parallel shuffle algorithm is to use a bijective function $f_k$ that for a given key $k$ defines a unique pseudo-random mapping on sets of indices $I_B \rightarrow I_X$. Assuming $I_B = \{0, \ldots B-1\}$ to describe indices in a minibatch and $I_X = \{0, \ldots, n-1\}$ indices in the data set, $f_k$ allows sampling a minibatch of elements from the input set in parallel without collisions. Previously, it has been proven that a Feistel network \cite{feistel1973cryptography} with more than two rounds is a pseudo-random bijective function, provided that it uses a round function that is pseudo-random and the set size is a power of two, i.e., $n=2^b$ \cite{luby1988construct}. 

In \cite{mitchell2021gpushuffling} this is generalised for arbitrary $n$ by taking the smallest bit-length $b$ such that $2^b > n$, applying the bijection on the index set of length $2^b$ and then removing all values larger than $n$ by an efficient GPU compaction algorithm. The overall runtime of this is in $\Theta(n)$.

However, for our application of sampling a minibatch, shuffling the entire index set is inefficient. Instead, we apply the Feistel network repeatedly on values from $I_B$ until all outputs are a value in $I_X$. Our generalised Feistel permutation generator can be given as
\begin{equation}
x_i^{(l)} = \left\{\begin{array}{ll}
  i & \text{, if $l = 0$} \\
  f_k(x_i^{(l-1)}) & \text{, if $x_i^{(l-1)} \geq n$ and $l > 0$} \\
  x_i^{(l-1)} & \text{, otherwise, }
\end{array}\right.
\label{eq:feistel}
\end{equation}
where $l$ is the iteration count and $i \in I_B$. Sampling a minibatch then requires $l B$ evaluations of the Feistel network for some factor $l$. These evaluations consist only of independent, parallel bitwise operations, making this approach very well-suited for GPU acceleration. 

The factor $l$ is the number of iterations that are required to converge Equation \ref{eq:feistel}, for which we will now establish a probabilistic upper bound. The probability with which the Feistel network $f_k$ returns an output $ f_k(x) < n$, given a bit-length $b$ such that $2^{b-1}< n < 2^b$ is
\begin{equation}
\begin{aligned}
  p = \Pr[f_k(x) < n] &= \frac{2^{b-1}+r}{2^b},
\end{aligned}
\end{equation}
where we let $r = n - 2^{b-1}$, i.e., the non-power-of-two residue of $n$.

Now, let $L_i$ be the random variable of the iteration count required to converge $x_i < n, \forall i \in I_B $ and $F_i = L_i - 1$ be the penultimate iteration count where some $x_i \geq n$. $F_i$ follows a negative binomial distribution, $Pr[F_i = f] = (1-p)^f p$. Its expected value is
\begin{equation}
\mathbb{E}\left[ F_i \right] = \frac{1-p}{p} = \frac{2^{b-1}-r}{2^{b-1}+r} ,
\end{equation}
and the cumulative distribution is
\begin{equation}
\Pr[F_i \leq f] = 1 - (1-p)^{f+1} = 1 - \left(\frac{2^{b-1}-r}{2^b}\right)^{f+1} .
\end{equation}
In the worst case with $r=1$, resulting in the largest gap to the next power of two, the expected values are 
\begin{equation}
\mathbb{E}[F_i] = \frac{2^{b-1}-1}{2^{b-1}+1}, \ \mathbb{E}[L_i] = \frac{2^b}{2^{b-1} + 1} .
\end{equation}
Hence with non-trivial data set sizes with $b \gg 1$, the expected number of iterations required to converge the permutation is $\mathbb{E}[L_i] \approx 2$, in the worst case. 
To estimate the maximum number of iterations for a given percentage $\theta$ of cases, we can calculate
\begin{equation}
\Pr[F_i \leq f_\theta] \leq \theta \Leftrightarrow f_\theta \leq \frac{\log(1-\theta)}{\log(2^{b-1}-r)-\log(2^b)} - 1 .
\end{equation}
Setting $\theta = 0.99$, $b \gg 1$ and the worst case $r=1$, we obtain
\begin{equation}
f_\theta \geq \frac{\log(1-\theta)}{\log(1/2))} - 1 \approx 5.65
\end{equation}
Therefore, in 99\% of cases we will see no more than six failures, i.e., seven total iterations (and no more than six in 95\% of cases). In practice, as long as the batch size is sufficiently small, precisely $q = B/n < 1/7$, our approach will require less evaluations of the bijection than the one of \cite{mitchell2021gpushuffling}.

\subsection{Privacy Bound and Hyperparameter Selection}
\label{sec:features}

\dppp offers an API to compute the perturbation hyperparameter $\sigma$ for the DP-VI algorithm via the $\texttt{approximate\_sigma}$ function to satisfy Requirement~\ref{req:dpsviParams} (\emph{Determine DP inference algorithm parameters automatically.}) This is realised by employing standard black-box optimisation techniques to find a suitable input such that the Fourier accountant arrives at the desired value for $\varepsilon$, given all other hyperparameters.

Finally, Requirement~\ref{req:privacyBoundsGuidance} (\emph{Provide assistance for the user in finding adequate privacy bounds}) is an important piece of guidance for users inexperienced with differential privacy. Optimal choice of $\varepsilon$ and $\delta$ depends on a delicate balance between the desired utility and the level of risk of privacy violation that is considered adequate by the user --- a choice, therefore, that can only be made by the user but requires knowledge of how $\varepsilon, \delta$ relate to concrete privacy risks. Unfortunately, this relationship is still an open research question, which makes it difficult to give detailed guidance to the user. Common practice is to require that $\delta < \frac{1}{N}$, where $N$ is the number of individual records in the data, and $\varepsilon \leq 1$, however \dppp does not enforce this currently to allow for free experimentation.

%%%%%%%%%%%%%%%%%%%%%%%% RESULTS %%%%%%%%%%%%%%%%%%%%%%%%%%%%%%%%%%%%%
\section{Evaluation}
\label{sec:results}

To demonstrate the performance and flexibility of our framework, we explore a few examples in this section. We first compare \dppp's runtime performance to TensorFlow on the implementation of a variational auto-encoder (Sec.~\ref{sec:TFComparison}). Afterwards, we show some results for the hierarchical logistic regression model discussed in detail in Sec.~\ref{sec:ComplexExample} and use that opportunity to highlight some privacy trade-offs in the regime of small data sets (Sec.~\ref{sec:LogRegModelEval}). Finally, we briefly compare results of \dppp to the original implementation of the DP-VI algorithm in \cite{jalko2016differentially} (Sec.~\ref{sec:GMMEval}) on a Gaussian mixture model.

\subsection{Comparison with TensorFlow Privacy}
\label{sec:TFComparison}

We first compare the performance of our \dppp package to an implementation of an identical model using the TensorFlow Probability framework \cite{tensorflow2017probability} with a manual implementation of the variational inference algorithm and privacy enabled by the TensorFlow Privacy package \cite{tensorflow2019privacy}. 

We choose a variational auto-encoder (VAE) model \cite{kingma2013auto} for this purpose. VAEs are generative models that consist of an encoder function, mapping data to parameters of a distribution on latent representations, and a decoder function, converting samples from the latent space to data samples. Generating data consists of drawing a sample from the distribution in the latent space and passing it through the decoder function. These mapping functions are represented by neural networks and therefore typically have a large number of parameters.

VAEs therefore provide an excellent test case for performance and have been previously used for the same purpose \cite{bingham2018pyro}. We use slight variations of the same model on different image classification data sets, namely MNIST \cite{lecun1998gradient}, Fashion-MNIST \cite{xiao2017fashionmnist} and CIFAR-10 \cite{krizhevsky2009learning}. For MNIST and Fashion-MNIST datasets, we use feed-forward networks with a single hidden layer encoder and decoder, consisting of \numprint{688884} trainable parameters in total. For CIFAR-10 dataset, we employ networks of 3 convolutional layers followed by a single dense layer, consisting of \numprint{640423} parameters in total.

Of particular interest are the respective runtime of the inference as well as a comparison of the inferred model to verify that \dppp is fast and accurate. Table~\ref{tab:performance_dppp_tfp} shows the runtime per epoch as well as the loss on the held-out test set after 20 \emph{epochs}, i.e., passes over the training data set of \numprint{60000} images (\numprint{50000} for CIFAR-10). We use a minibatch size of 128. The reported numbers are averages over 20 training processes that were run using a single NVIDIA Tesla V100 32G GPU. All runs used the Adam optimiser \cite{kingma2015adam} for parameter updates after computing gradients. For the DP variants we used $\sigma = 1.5$, resulting in $\varepsilon \approx 0.5$ for $\delta = 1/60000$ ($\delta = 1/50000$ for CIFAR).

\begin{table*}
    \footnotesize
    \centering
    \setlength\tabcolsep{5pt}
    \begin{tabular}{r | c | r r | r r }
      \toprule % from booktabs package
      \bfseries & & \multicolumn{2}{c|}{DP-VI} & \multicolumn{2}{c}{Non-private VI} \\
      \bfseries Data Set & Framework & Wall Time [s] & Final Loss & Wall Time [s] & Final Loss \\
      \midrule 
      MNIST & \dppp & $0.56 \pm 0.00$ & $174.06 \pm 0.59$ & $0.34 \pm 0.00$ & $99.63 \pm \phantom{0}0.30$ \\
      & TF & $6.43 \pm 0.17$ & $171.26 \pm 3.67$ & $1.58 \pm 0.05$ & $105.50 \pm \phantom{0}1.84$ \\
      \hline
       Fashion & \dppp & $0.55 \pm 0.01$ & $304.74 \pm 0.68$ & $0.33 \pm 0.00$ & $243.99 \pm \phantom{0}0.29$ \\
       & TF & $7.19 \pm 0.14$ & $303.14 \pm 8.79$ & $1.70 \pm 0.06$ & $244.40 \pm \phantom{0}7.56$ \\
       \hline
       CIFAR & \dppp & $4.22 \pm 0.01$ & $2123.05 \pm 0.25$ & $1.64 \pm 0.01$ & $1903.29 \pm \phantom{0}2.14$ \\
       & TF & $49.34 \pm 0.07$ & $2129.20 \pm 0.26$ & $2.23 \pm 0.05$ & $2038.92 \pm 11.06$ \\
      \bottomrule % from booktabs package
    \end{tabular}
    \caption{Performance comparison of \dppp against TensorFlow probability with TensorFlow Privacy. \dppp achieves significant speed-up over TensorFlow Privacy with similar loss. Values shown in the left half of the table are runtime per epoch as well as the final loss (negative ELBO, lower is better) value on test set after 20 epochs for the differentially-private VI in both frameworks, using the Adam optimiser for parameter updates in all cases. The right part of the table shows the same results for non-private inference, where NumPyro takes the place of \dppp.}\label{tab:performance_dppp_tfp}
\end{table*}

From the comparison, we can see that the end losses with MNIST and Fashion-MNIST data sets are comparable across frameworks, and all cases were observed to converge well. However, we found CIFAR-10 data set to be more challenging to train due to large variation in samples, relatively modest number of samples per class and three colour channels. Neither framework was able to learn good representations of CIFAR-10 with DP for our choice of hyperparameters. The aforementioned challenges may also explain the higher discrepancy between the non-DP losses. In terms of performance, \dppp consistently outperforms the TensorFlow (TF) implementation by a factor of $\sim$10 for all data sets. Additionally, the relative performance loss of DP-VI compared to non-private inference in the same framework is lower in \dppp (up to $\sim2.5$-fold for CIFAR-10) compared to the TF implementation (up to $\sim22$-fold).

For \dppp, we also compare using the Feistel-based GPU-optimised minibatch sampler against using JAX's built-in \texttt{jax.random.choice} method and summarise the results in Table~\ref{tab:performanc_minibatch_sampling}. We observe that our optimised sampler consistently yields a $\sim 100$ ms speed-up ($15\%$ on MNIST) per epoch on all data sets. This similarity is due to the similar sizes of the data sets resulting in similar amounts of total iterations and thus invocations of the subsampling.

\begin{table}
    \footnotesize
    \centering
    \setlength\tabcolsep{5pt}
    \begin{tabular}{r | c | r r}
      \toprule % from booktabs package
      \bfseries Data Set & Sampler & Wall Time [s] & Final Loss \\
      \midrule 
      MNIST & Feistel & $0.56 \pm 0.00$ & $174.06 \pm 0.59$ \\
      & Built-in & $0.66 \pm 0.01$ & $174.04 \pm 0.61$ \\
      \hline
       Fashion & Feistel & $0.55 \pm 0.01$ & $304.74 \pm 0.68$ \\
       & Built-in & $0.65 \pm 0.01$ & $304.75 \pm 0.61$ \\
       \hline
       CIFAR & Feistel & $4.22 \pm 0.01$ & $2123.05 \pm 0.25$ \\
       & Built-in & $4.31 \pm 0.01$ & $2123.05 \pm 0.018$ \\
      \bottomrule % from booktabs package
    \end{tabular}
    \caption{Performance comparison of the \dppp Feistel-based minibatch sampler compared to sampling based on JAX default routines.}\label{tab:performanc_minibatch_sampling}
\end{table}

\subsection{Effect of Batch Size}
\begin{figure}[bht]
    \centering
    \includegraphics[width=3in]{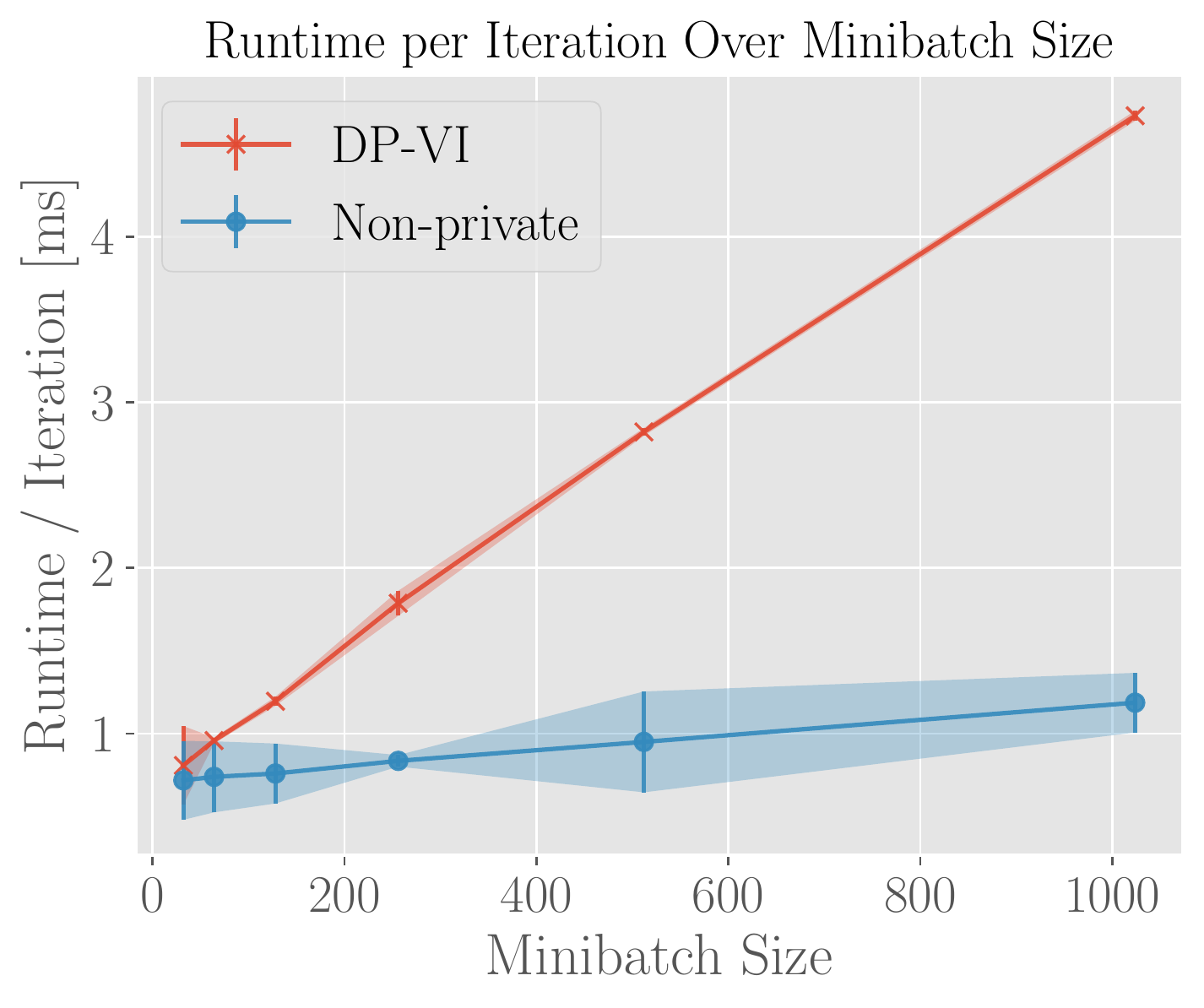}
    \caption{Influence of minibatch size on runtime performance with and without differential privacy. Results are averages over \numprint{100} iterations. Error bars and shaded area indicate standard error.}
    \label{fig:performance_minibatches}
\end{figure}

While the vectorised implementation of the per-instance gradient computation and manipulation eliminates much of the time overhead required in the DP-VI algorithm, it does not remove it completely. This is due to the additional steps in DP-VI, such as gradient clipping, summing and perturbing, but may also be an effect of a larger memory footprint due to holding gradient values for each data instance intermittently. This necessitates a larger amount of memory accesses, which can slow down computation. We therefore investigate the effect of minibatch size on the runtime of the DP-VI implementation in \dppp for MNIST. In Figure~\ref{fig:performance_minibatches} we plot the average runtime per iteration (over \numprint{100} iterations) over the size of minibatches for DP-VI in \dppp and non-private inference in NumPyro. We see that, as expected, runtime per iteration increases more steeply for DP-VI than the non-private case. However, we also observe that DP-VI runtime increases only linearly with minibatch size, which is in line with our expectations.

\subsection{Hierarchical Logistic Regression and Small Data}
\label{sec:LogRegModelEval}

We next evaluate the implementation of the hierarchical logistic regression model presented in Section~\ref{sec:ComplexExample} with small data sets. These are generally problematic in privacy-preserving inference because every single data point has a comparatively larger effect on the outcome. Achieving good model performance therefore requires looser privacy bounds than in the case with larger data sets. \dppp makes it easy for the user explore the trade-off between privacy and utility via its fast inference and the ease of adjusting the algorithm by simply specifying the privacy parameters from which the perturbation noise is automatically derived.

\begin{figure}[t!]
    \centering
    \includegraphics[width=3in]{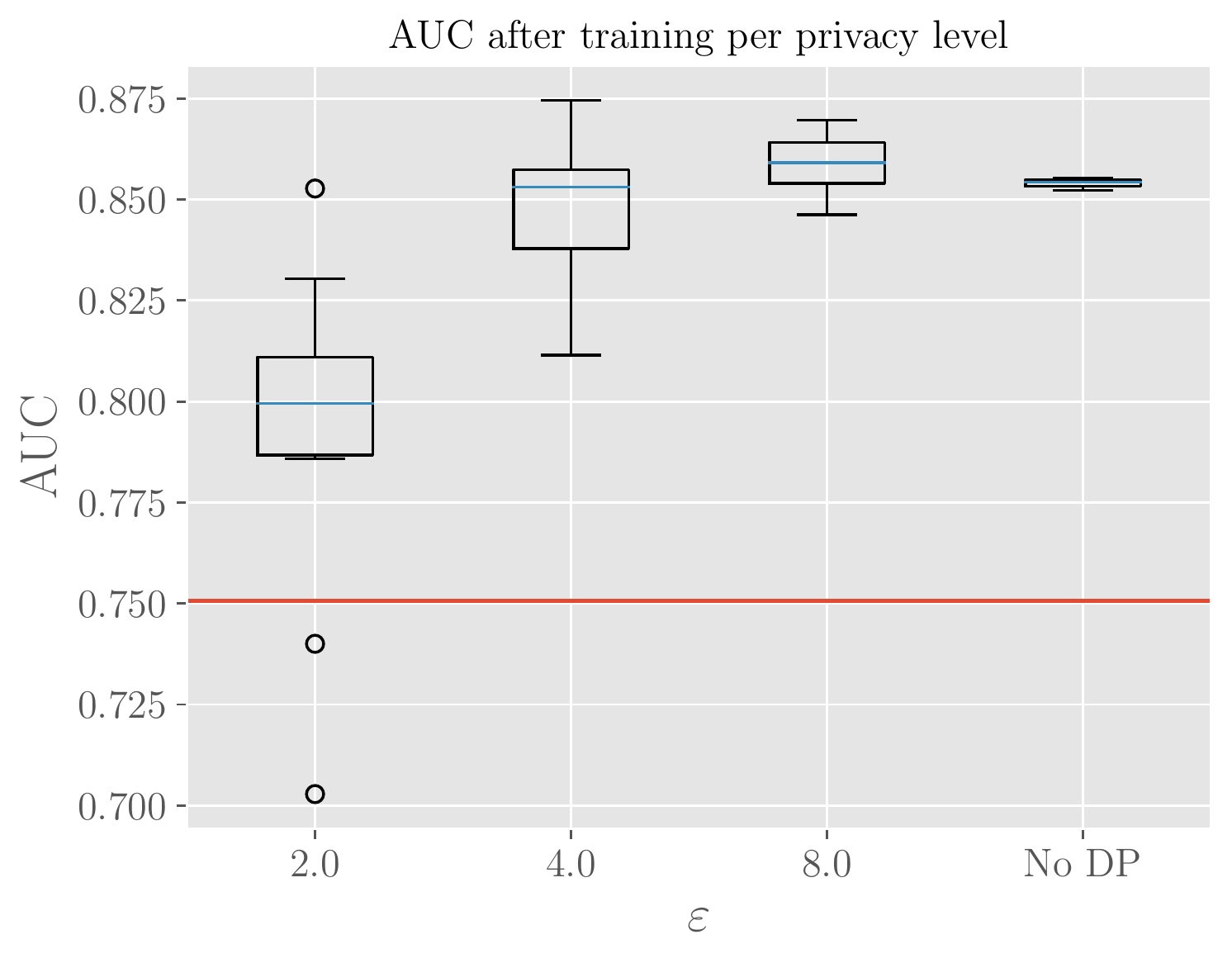}
    \caption{AUC of the hierarchical logistic regression model after training on 500 data points. Ten runs of \numprint{100000} iterations each were performed for each level of privacy and a non-private run, the resulting distribution of AUC values are shown as box plots. The red line indicates the AUC of a non-private non-hierarchical logistic regression model as a simple baseline.}
    \label{fig:HLR_n500_per_eps}
\end{figure}

In the following experiments we use synthetic training data sets of varying size $N$ that follow the hierarchical structure described in Section~\ref{sec:ComplexExample}. Each data point is five-dimensional and points are split into $L=3$ groups that are in turn described by $K=3$ variables each. To evaluate the goodness-of-fit of the trained model we evaluate the area under the ROC curve (AUC) on a held-out test set of the same size $N$ as the training set.

Figure~\ref{fig:HLR_n500_per_eps} shows the AUC after training on a data set with $N=500$ data points for \numprint{100000} iterations. Results are shown for different levels of privacy as well as non-private variational inference for ten runs with different random seeds. Privacy bound $\delta$ was kept fixed to $\frac{1}{N}$. Each run took less than 15 seconds on a commodity laptop without GPU acceleration. The red line is the AUC for non-privately fitting a simple (non-hierarchical) logistic regression model using scikit-learn \cite{scikitlearn2011}. The trade-off between privacy and utility is clearly visible: Smaller values of $\varepsilon$ corresponding to stricter privacy constraints lead to lower AUC on average and a larger spread of results over different runs. Runs for $\varepsilon = 4$ or larger are close to the AUC of the non-private model on average but exhibit larger spread. Runs for $\varepsilon = 2$ fall short of this but still outperform the simpler baseline. This highlights a strength of the probabilistic modelling approach particularly relevant for privacy-preserving machine learning: By encoding prior knowledge of the generative process underlying the data in a principled way, privacy budget does not need to be spent to learn this a-priori known structure, allowing more capacity to learn the remaining parameters of the model.

\begin{figure}[t!]
    \centering
    \includegraphics[width=3in]{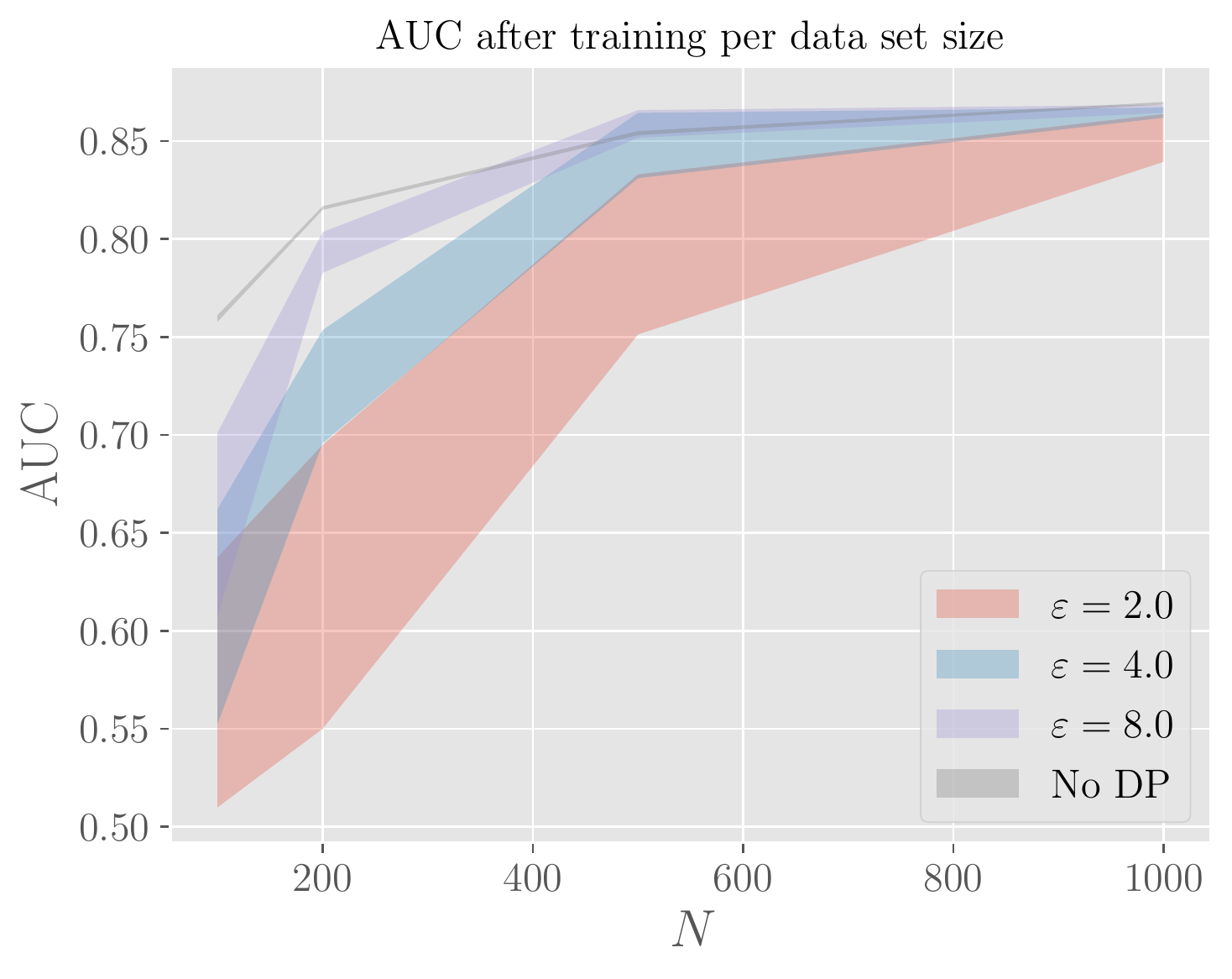}
    \caption{AUC of the hierarchical logistic regression model after training on data sets of different sizes (100, 200, 500, 1000) for different privacy levels and non-privately. The model was trained 10 times for \numprint{100000} iterations for each data set size and privacy bound. The graphs show the area within one standard deviation above and below the mean of the resulting distribution of AUC values.}
    \label{fig:HLR_i100000_per_n_per_eps}
\end{figure}

Figure~\ref{fig:HLR_i100000_per_n_per_eps} shows the effect of data set size for the same privacy levels after training for \numprint{100000} iterations. The graphs show the spread of one standard deviation above and below the mean over ten runs for each data set size and privacy level. Smaller data sets result in lower average AUC and larger spread. For data sets of less than 500 records utility deteriorates rapidly.

\begin{figure}[t]
    \centering
    \includegraphics[width=3.3in]{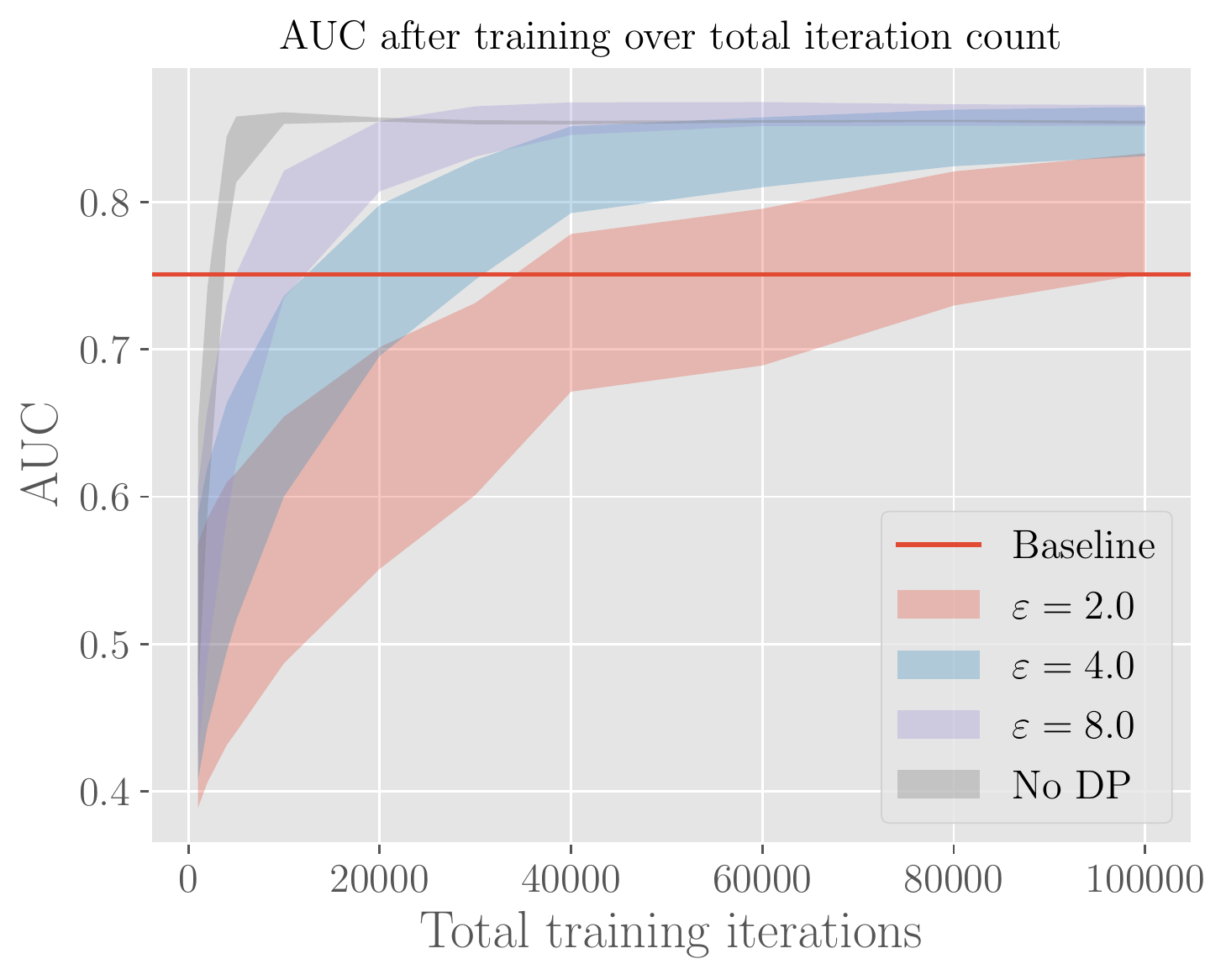}
    \caption{AUC of the hierarchical logistic regression model after training on $N=500$ data points for different amounts of training iterations. Plotted as in Figure~\ref{fig:HLR_i100000_per_n_per_eps}. The additional horizontal line shows the AUC of the simple non-private non-hierarchical logistic regression baseline.}
    \label{fig:HLR_n500_total_iters}
\end{figure}

We finally explore the effects of the number of training iterations. Since the amount of noise added to perturb gradients in the DP-VI algorithm increases with the number of iterations, one could expect that choosing too large a number of iterations will negatively effect the learning. Figure~\ref{fig:HLR_n500_total_iters} shows the evolution of AUC for $N=500$ for different numbers of total training iterations. Note that these are results of separate runs each with a different number of iterations and thus different amounts of perturbation per iteration, \emph{not} the evolution of results over a single long training run.

We observe clearly that due to the privacy perturbations of gradients during training, the DP-VI algorithm takes longer to converge than non-private variational inference. Stricter privacy bounds move convergence to higher iteration counts. Contrary to expectation, despite the larger perturbations required for larger iteration counts, we see a general trend in improved utility for longer training (for $N=100$, this trend  continues up to \numprint{500000} iterations). In our experiments we observe no negative impact of increasing the iteration count on the final AUC even if the training converges earlier, indicating that the DP-VI algorithm is very robust to the privacy perturbations.

\subsection{Gaussian Mixture Model}
\label{sec:GMMEval}

We further demonstrate the ease of specifying expressive models in \dppp by replicating an experiment on a Gaussian mixture model from the original DP-VI paper \cite{jalko2016differentially}. They used two-dimensional data generated from 5 clusters of spherical Gaussians and trained the model for \numprint{1000} iterations for different levels of privacy. The evaluation is in terms of log-likelihood of a held-out test set on the learned predictive model.

The original code\footnote{Available at \url{https://github.com/DPBayes/DPVI-code/}} required model specific implementation of the DP-VI algorithm due to the absence of a generic framework for privacy-preserving probabilistic modelling. Using \dppp it suffices to simply write out the model and an implementation of log-probability calculations and sampling routines for a Gaussian mixture distribution as an implementation of NumPyro's \texttt{Distribution} class. These are shown in Appendix~\ref{app:GMMCode}.

Figure~\ref{fig:GMMResults} shows the resulting log-likelihood (higher is better) of the test data after training the model using the \dppp implementation and the original code of \cite{jalko2016differentially}. Shown are the average with standard error over five runs in both cases. To keep results comparable we use the levels of privacy perturbation (parameter $\sigma$) from the original code for \dppp. The results from \dppp are very consistent with to slightly better than those from the original implementation but exhibit much less variability due to randomness in the inference algorithm, which is particularly pronounced for $\varepsilon = 0.1$ in the original paper code.

\begin{figure}
    \centering
    \includegraphics[width=3in]{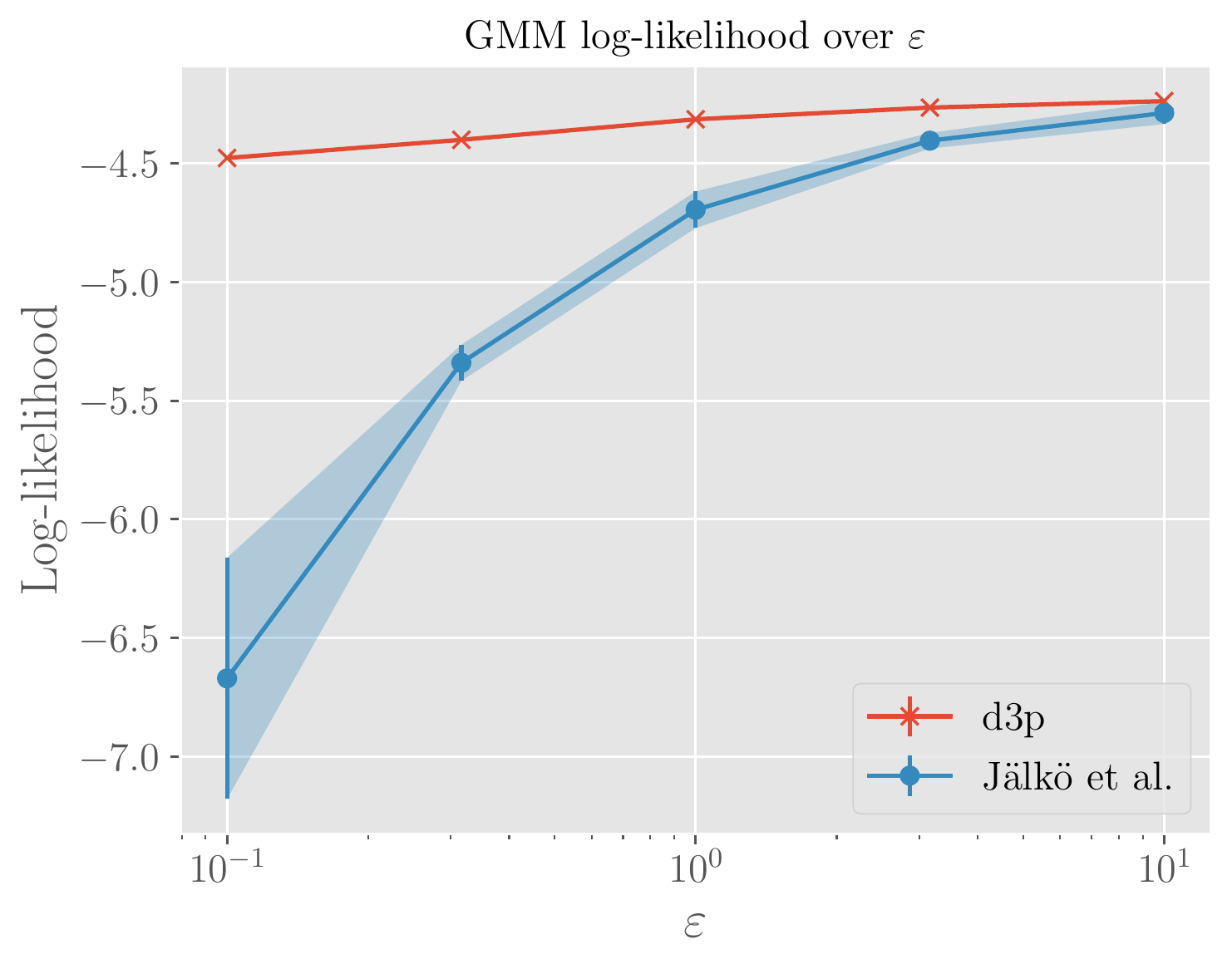}
    \caption{Log-likelihood for test data of the Gaussian mixture model implemented with \dppp and the DP-VI implementation of \citeauthor{jalko2016differentially}\cite{jalko2016differentially}. Graphs show average over five runs with error bars and shaded area indicating standard error (negligible for \dppp).}
    \label{fig:GMMResults}
\end{figure}

%%%%%%%%%%%%%%%%%%%%%%%%% RELATED WORK %%%%%%%%%%%%%%%%%%%%%%%%%%%%%%%%5
\section{Related Work}

Major popular probabilistic programming frameworks for the Python programming language are Edward2 \cite{tran2018simple}, TensorFlow Probability \cite{tensorflow2017probability}, (Num)Pyro \cite{bingham2018pyro, phan2019composable} , PyMC \cite{fonnesbeck2015pymc}, and Stan \cite{carpenter2017stan}. Edward2, TensorFlow Probability, Pyro and PyMC allow the user to declare models and run the inference from the Python programming language and differ mainly in the computation framework they rely on to run the inference (Edward2 and TensorFlow Probability use TensorFlow \cite{tensorflow2015whitepaper}, Pyro uses PyTorch \cite{pytorch2019framework} and PyMC uses Theano \cite{theano2016framework}; NumPyro is a direct port of Pyro to JAX). The Stan framework follows a different approach and requires models to be specified in a dedicated domain-specific language, which is then evaluated using the Stan runtime, which can be invoked from Python or other major programming languages. None of these frameworks currently offers support for privacy-preserving inference.

A number of general implementations for differentially private machine learning exist for the popular frameworks. Notable are TensorFlow Privacy \cite{tensorflow2019privacy} for TensorFlow, Opacus \cite{facebook2020opacus} and PyVacy \cite{waites2019pyvacy} for PyTorch. These generally provide implementations of the DP-SGD algorithm as alternative optimisers for the computational framework. In principle they could be combined with the dominant probabilistic programming framework for the respective backend, however this integration is usually not as seamless as one would desire. These implementations of DP also often suffer from poor performance in the implementation that can usually be traced back to inefficient computation of per-instance gradients \cite{subramani2020enabling}. With \dppp we aim to provide better integration with high performance by directly targeting the NumPyro probabilistic programming framework. We are not aware of any general library of DP-SGD for the JAX framework that could be used with NumPyro to achieve the same goal. As far as we are aware, none of these libraries takes active measures to address the previously discussed technical issues of implementing differential privacy.

%%%%%%%%%%%%%%%%%% CONCLUSION %%%%%%%%%%%%%%%%%%%%%%%%%%%%%%%
\section{Conclusion}
We have presented our \dppp package which extends the NumPyro probabilistic programming framework with runtime efficient differentially private inference. We demonstrated the use of our framework and the expressiveness of the probabilistic programming approach on an extensive example and highlighted the requirements and corresponding implementation choices for our software. Our goal is to provide a helpful tool that encourages use of probabilistic programming as a viable approach to modelling data for privacy practitioners, as well as lowers the threshold for adoption of privacy-preserving methods for probabilistic modelling experts.

For future work our main focus is addressing the remaining technical implementation issues of differential privacy in real computer systems, namely predictable random number generation and finite-precision number representation (cf.~Sec.~\ref{sec:PracticalDPBackground}), as the main obstacle for deployment in production settings. Promising solutions for this are (1) use of a cryptographically secure random number generator (CSPRNG) for DP perturbation and minibatch sampling and (2) adoption of the discrete Gaussian mechanism \cite{canonne2020discrete, kairouz2021discrete}.

\section*{Acknowledgements}
This work was supported by the Academy of Finland (Flagship programme: Finnish Center for Artificial Intelligence FCAI, and grants 292334, 319264, 325572 and 325573) and by the Strategic Research Council at the Academy of Finland (grant 336032). We also acknowledge the computational resources provided by the Aalto Science-IT Project.

%%% references
% \printbibliography
\bibliographystyle{plainnat}
\bibliography{references}

\begin{thebibliography}{43}
\providecommand{\natexlab}[1]{#1}
\providecommand{\url}[1]{\texttt{#1}}
\expandafter\ifx\csname urlstyle\endcsname\relax
  \providecommand{\doi}[1]{doi: #1}\else
  \providecommand{\doi}{doi: \begingroup \urlstyle{rm}\Url}\fi

\bibitem[Abadi et~al.(2015)]{tensorflow2015whitepaper}
Mart\'{\i}n Abadi et~al.
\newblock {TensorFlow}: Large-scale machine learning on heterogeneous systems,
  2015.
\newblock URL \url{https://www.tensorflow.org/}.
\newblock Software available from tensorflow.org.

\bibitem[Abadi et~al.(2016)]{abadi2016dpsgd}
Martin Abadi et~al.
\newblock Deep learning with differential privacy.
\newblock In \emph{Proceedings of the 2016 ACM SIGSAC conference on computer
  and communications security}, pages 308--318, 2016.

\bibitem[Bingham et~al.(2018)]{bingham2018pyro}
Eli Bingham et~al.
\newblock {Pyro: Deep Universal Probabilistic Programming}.
\newblock \emph{{arXiv} preprint {arXiv}:1810.09538}, 2018.

\bibitem[Bradbury et~al.(2018)Bradbury, Frostig, Hawkins, Johnson, Leary,
  Maclaurin, and Wanderman-Milne]{bradbury2018jax}
James Bradbury, Roy Frostig, Peter Hawkins, Matthew~James Johnson, Chris Leary,
  Dougal Maclaurin, and Skye Wanderman-Milne.
\newblock {JAX}: composable transformations of {P}ython+{N}um{P}y programs.
\newblock \url{https://github.com/google/jax}, 2018.

\bibitem[Canonne et~al.(2020)Canonne, Kamath, and Steinke]{canonne2020discrete}
Cl\'{e}ment~L Canonne, Gautam Kamath, and Thomas Steinke.
\newblock The discrete gaussian for differential privacy.
\newblock In H.~Larochelle, M.~Ranzato, R.~Hadsell, M.~F. Balcan, and H.~Lin,
  editors, \emph{Advances in Neural Information Processing Systems}, volume~33,
  pages 15676--15688. Curran Associates, Inc., 2020.

\bibitem[Carpenter et~al.(2017)]{carpenter2017stan}
Bob Carpenter et~al.
\newblock Stan: a probabilistic programming language.
\newblock \emph{Journal of Statistical Software}, 76\penalty0 (1), 2017.

\bibitem[Dillon et~al.(2017)]{tensorflow2017probability}
Joshua~V. Dillon et~al.
\newblock Tensorflow distributions.
\newblock \emph{{arXiv} preprint {arXiv}:1711.10604}, 2017.

\bibitem[Dwork and Roth(2014)]{dwork2014algorithmic}
Cynthia Dwork and Aaron Roth.
\newblock The algorithmic foundations of differential privacy.
\newblock \emph{Foundations and Trends in Theoretical Computer Science},
  9\penalty0 (3-4):\penalty0 211--407, 2014.

\bibitem[Dwork et~al.(2006{\natexlab{a}})Dwork, Kenthapadi, McSherry, Mironov,
  and Naor]{dwork2006our}
Cynthia Dwork, Krishnaram Kenthapadi, Frank McSherry, Ilya Mironov, and Moni
  Naor.
\newblock Our data, ourselves: Privacy via distributed noise generation.
\newblock In \emph{Annual International Conference on the Theory and
  Applications of Cryptographic Techniques}, pages 486--503. Springer,
  2006{\natexlab{a}}.

\bibitem[Dwork et~al.(2006{\natexlab{b}})Dwork, McSherry, Nissim, and
  Smith]{dwork2006calibrating}
Cynthia Dwork, Frank McSherry, Kobbi Nissim, and Adam Smith.
\newblock Calibrating noise to sensitivity in private data analysis.
\newblock In \emph{Theory of cryptography conference}, pages 265--284.
  Springer, 2006{\natexlab{b}}.

\bibitem[Erlingsson et~al.(2019)Erlingsson, Mironov, Raghunathan, and
  Song]{erlingsson2019that}
Úlfar Erlingsson, Ilya Mironov, Ananth Raghunathan, and Shuang Song.
\newblock That which we call private.
\newblock \emph{{arXiv} preprint {arXiv}:1908.03566}, 2019.

\bibitem[Facebook(2020)]{facebook2020opacus}
Facebook.
\newblock Opacus.
\newblock \url{https://opacus.ai/}, 2020.

\bibitem[Feistel(1973)]{feistel1973cryptography}
Horst Feistel.
\newblock Cryptography and computer privacy.
\newblock \emph{Scientific american}, 228\penalty0 (5):\penalty0 15--23, 1973.

\bibitem[Fonnesbeck et~al.(2015)Fonnesbeck, Patil, Huard, and
  Salvatier]{fonnesbeck2015pymc}
Chris Fonnesbeck, Anand Patil, David Huard, and John Salvatier.
\newblock {PyMC}: Bayesian stochastic modelling in python.
\newblock \emph{Astrophysics Source Code Library}, 2015.

\bibitem[Frostig et~al.(2018)Frostig, Johnson, and Leary]{frostig2018compiling}
Roy Frostig, Matthew~James Johnson, and Chris Leary.
\newblock Compiling machine learning programs via high-level tracing.
\newblock \emph{Systems for Machine Learning}, 2018.

\bibitem[Garfinkel and Leclerc(2020)]{garfinkel2020randomness}
Simson~L. Garfinkel and Philip Leclerc.
\newblock Randomness concerns when deploying differential privacy.
\newblock In \emph{Proceedings of the 19th Workshop on Privacy in the
  Electronic Society}, WPES'20, page 73–86, New York, NY, USA, 2020.
  Association for Computing Machinery.
\newblock ISBN 9781450380867.
\newblock \doi{10.1145/3411497.3420211}.

\bibitem[Harris et~al.(2020)]{harris2020array}
Charles~R. Harris et~al.
\newblock Array programming with {{N}um{P}y}.
\newblock \emph{Nature}, 585\penalty0 (7825):\penalty0 357--362, September
  2020.
\newblock \doi{10.1038/s41586-020-2649-2}.

\bibitem[J{\"a}lk{\"o} et~al.(2017)J{\"a}lk{\"o}, Dikmen, and
  Honkela]{jalko2016differentially}
Joonas J{\"a}lk{\"o}, Onur Dikmen, and Antti Honkela.
\newblock Differentially private variational inference for non-conjugate
  models.
\newblock In \emph{Uncertainty in Artificial Intelligence 2017 Proceedings of
  the 33rd Conference, UAI 2017}. The Association for Uncertainty in Artificial
  Intelligence, 2017.

\bibitem[J{\"a}lk{\"o} et~al.(2021)J{\"a}lk{\"o}, Lagerspetz, Haukka, Tarkoma,
  Honkela, and Kaski]{jalko2019privacy}
Joonas J{\"a}lk{\"o}, Eemil Lagerspetz, Jari Haukka, Sasu Tarkoma, Antti
  Honkela, and Samuel Kaski.
\newblock Privacy-preserving data sharing via probabilistic modeling.
\newblock \emph{Patterns}, 2\penalty0 (7):\penalty0 100271, 2021.
\newblock ISSN 2666-3899.
\newblock \doi{10.1016/j.patter.2021.100271}.

\bibitem[Jordan et~al.(1999)Jordan, Ghahramani, Jaakkola, and
  Saul]{jordan1999introduction}
Michael~I. Jordan, Zoubin Ghahramani, Tommi~S. Jaakkola, and Lawrence~K. Saul.
\newblock An introduction to variational methods for graphical models.
\newblock \emph{Machine learning}, 37\penalty0 (2):\penalty0 183--233, 1999.

\bibitem[Kairouz et~al.(2021)Kairouz, Liu, and Steinke]{kairouz2021discrete}
Peter Kairouz, Ziyu Liu, and Thomas Steinke.
\newblock The distributed discrete gaussian mechanism for federated learning
  with secure aggregation.
\newblock In Marina Meila and Tong Zhang, editors, \emph{Proceedings of the
  38th International Conference on Machine Learning}, volume 139 of
  \emph{Proceedings of Machine Learning Research}, pages 5201--5212. PMLR,
  18--24 Jul 2021.

\bibitem[Kingma and Ba(2015)]{kingma2015adam}
Diederik~P. Kingma and Jimmy Ba.
\newblock Adam: A method for stochastic optimization.
\newblock In \emph{International Conference on Learning Representations (ICLR
  2015)}, 2015.

\bibitem[Kingma and Welling(2014)]{kingma2013auto}
Diederik~P. Kingma and Max Welling.
\newblock Auto-encoding variational {B}ayes.
\newblock In \emph{2nd International Conference on Learning Representations
  (ICLR 2014)}, 2014.

\bibitem[Koskela et~al.(2020)Koskela, J{\"a}lk{\"o}, and
  Honkela]{koskela2020computing}
Antti Koskela, Joonas J{\"a}lk{\"o}, and Antti Honkela.
\newblock Computing tight differential privacy guarantees using {FFT}.
\newblock In \emph{International Conference on Artificial Intelligence and
  Statistics}, pages 2560--2569. PMLR, 2020.

\bibitem[Krizhevsky(2009)]{krizhevsky2009learning}
Alex Krizhevsky.
\newblock Learning multiple layers of features from tiny images.
\newblock Technical report, University of Toronto, 2009.

\bibitem[LeCun et~al.(1998)LeCun, Bottou, Bengio, and
  Haffner]{lecun1998gradient}
Yann LeCun, L{\'e}on Bottou, Yoshua Bengio, and Patrick Haffner.
\newblock Gradient-based learning applied to document recognition.
\newblock \emph{Proceedings of the IEEE}, 86\penalty0 (11):\penalty0
  2278--2324, 1998.

\bibitem[Luby and Rackoff(1988)]{luby1988construct}
Michael Luby and Charles Rackoff.
\newblock How to construct pseudorandom permutations from pseudorandom
  functions.
\newblock \emph{SIAM Journal on Computing}, 17\penalty0 (2):\penalty0 373--386,
  1988.

\bibitem[Mironov et~al.(2009)Mironov, Pandey, Reingold, and
  Vadhan]{mironov2009computational}
Ilya Mironov, Omkant Pandey, Omer Reingold, and Salil Vadhan.
\newblock Computational differential privacy.
\newblock In \emph{Annual International Cryptology Conference}, pages 126--142.
  Springer, 2009.

\bibitem[Mitchell et~al.(2021)Mitchell, Stokes, Frank, and
  Holmes]{mitchell2021gpushuffling}
Rory Mitchell, Daniel Stokes, Eibe Frank, and Geoffrey Holmes.
\newblock Bandwidth-optimal random shuffling for {GPU}s.
\newblock \emph{{arXiv} preprint {arXiv}:2106.06161}, abs/2106.06161, 2021.

\bibitem[Paszke et~al.(2019)]{pytorch2019framework}
Adam Paszke et~al.
\newblock {PyTorch}: An imperative style, high-performance deep learning
  library.
\newblock In \emph{Advances in Neural Information Processing Systems 32}, pages
  8024--8035. Curran Associates, Inc., 2019.

\bibitem[Pedregosa et~al.(2011)Pedregosa, Varoquaux, Gramfort, Michel, Thirion,
  Grisel, Blondel, Prettenhofer, Weiss, Dubourg, Vanderplas, Passos,
  Cournapeau, Brucher, Perrot, and Duchesnay]{scikitlearn2011}
F.~Pedregosa, G.~Varoquaux, A.~Gramfort, V.~Michel, B.~Thirion, O.~Grisel,
  M.~Blondel, P.~Prettenhofer, R.~Weiss, V.~Dubourg, J.~Vanderplas, A.~Passos,
  D.~Cournapeau, M.~Brucher, M.~Perrot, and E.~Duchesnay.
\newblock Scikit-learn: Machine learning in {P}ython.
\newblock \emph{Journal of Machine Learning Research}, 12:\penalty0 2825--2830,
  2011.

\bibitem[Phan et~al.(2019)Phan, Pradhan, and Jankowiak]{phan2019composable}
Du~Phan, Neeraj Pradhan, and Martin Jankowiak.
\newblock Composable effects for flexible and accelerated probabilistic
  programming in {N}um{P}yro.
\newblock \emph{{arXiv} preprint {arXiv}:1912.11554}, 2019.

\bibitem[Radebaugh and Erlingsson(2019)]{tensorflow2019privacy}
Carey Radebaugh and Ulfar Erlingsson.
\newblock Introducing {TensorFlow} privacy: Learning with differential privacy
  for training data.
\newblock TensorFlow Blog,
  \url{https://blog.tensorflow.org/2019/03/introducing-tensorflow-privacy-learning.html},
  2019.

\bibitem[Song et~al.(2013)Song, Chaudhuri, and Sarwate]{song2013stochastic}
Shuang Song, Kamalika Chaudhuri, and Anand~D. Sarwate.
\newblock Stochastic gradient descent with differentially private updates.
\newblock In \emph{2013 IEEE Global Conference on Signal and Information
  Processing}, pages 245--248. IEEE, 2013.

\bibitem[Stokes and Mitchell(2021)]{stokesfeistel}
Daniel Stokes and Rory Mitchell.
\newblock {CUDA-Shuffle}: {GPU} shuffle using bijective functions.
\newblock \url{https://github.com/djns99/CUDA-Shuffle}, 2021.

\bibitem[Subramani et~al.(2020)Subramani, Vadivelu, and
  Kamath]{subramani2020enabling}
Pranav Subramani, Nicholas Vadivelu, and Gautam Kamath.
\newblock Enabling fast differentially private {SGD} via just-in-time
  compilation and vectorization.
\newblock \emph{{arXiv} preprint {arXiv}:2010.09063}, 2020.

\bibitem[{Theano Development Team}(2016)]{theano2016framework}
{Theano Development Team}.
\newblock Theano: A python framework for fast computation of mathematical
  expressions.
\newblock \emph{{arXiv} preprint {arXiv}:1605.02688}, 2016.

\bibitem[Titsias and L{\'a}zaro-Gredilla(2014)]{titsias2014doubly}
Michalis Titsias and Miguel L{\'a}zaro-Gredilla.
\newblock Doubly stochastic variational {B}ayes for non-conjugate inference.
\newblock In \emph{International conference on machine learning}, pages
  1971--1979, 2014.

\bibitem[Tran et~al.(2018)]{tran2018simple}
Dustin Tran et~al.
\newblock Simple, distributed, and accelerated probabilistic programming.
\newblock In \emph{Neural Information Processing Systems}, 2018.

\bibitem[Wainwright and Jordan(2008)]{wainwright2008graphical}
Martin~J. Wainwright and Michael~Irwin Jordan.
\newblock \emph{Graphical models, exponential families, and variational
  inference}.
\newblock Now Publishers Inc, 2008.

\bibitem[Waites(2019)]{waites2019pyvacy}
Chris Waites.
\newblock {PyVacy}.
\newblock \url{https://github.com/ChrisWaites/pyvacy}, 2019.

\bibitem[Wong and Mason(1985)]{wong1985hierarchical}
George~Y. Wong and William~M. Mason.
\newblock The hierarchical logistic regression model for multilevel analysis.
\newblock \emph{Journal of the American Statistical Association}, 80\penalty0
  (391):\penalty0 513--524, 1985.
\newblock ISSN 01621459.

\bibitem[Xiao et~al.(2017)Xiao, Rasul, and Vollgraf]{xiao2017fashionmnist}
Han Xiao, Kashif Rasul, and Roland Vollgraf.
\newblock Fashion-mnist: a novel image dataset for benchmarking machine
  learning algorithms.
\newblock \emph{{arXiv} preprint {arXiv}:1708.07747}, 2017.

\end{thebibliography}

\appendix
\section{Code for the Gaussian Mixture Model}
\label{app:GMMCode}

We present the implement for the Gaussian mixture model (GMM) used in the experiment in Section~\ref{sec:GMMEval} in Listings~\ref{lst:gmm_dist} and \ref{lst:gmm_model}.

Mathematically a GMM can be specified as
\begin{align*}
    p(\bm{x}_i | z_i) &= \mathcal{N}\left(\bm{x}_i; \bm{\mu}_{z_i}, \bm{\Sigma}_{z_i}\right), \nonumber \\
    p(z_i) &= \text{Categorical}\left( \pi_1, \ldots, \pi_K \right)
\end{align*}
where $z_i \in \{1, \ldots, K\}$ is a latent variable that indicates the mixture component that sample $\bm{x}_i$. Conditioned on $z_i$, $\bm{x}_i$ follows a regular Normal distribution. The categorical probabilities $\pi_j$ and the parameters $\bm{\mu}_j$, $\bm{\Sigma}_j$ of the mixture components are the parameters of the GMM.

Sampling from the GMM can therefore be implemented by first sampling $z_i$ from the Categorical distribution, then sampling $\bm{x}_i$ from the Normal distribution indicated by $z_i$. This is presented in the \texttt{sample} function of the \texttt{GaussianMixtureModel} class in Listing~\ref{lst:gmm_dist}. 

Subclassing \texttt{Distribution} enables us to provide a method for computing the log-probability of the Gaussian mixture where we marginalise out the latent variables to avoid these issues following \cite{jalko2016differentially}:
\begin{equation*}
    \log p(\bm{x}_i) = \log\sum_{j=1}^K \left(\pi_j \mathcal{N}\left(\bm{x}_i; \bm{\mu}_j, \bm{\Sigma}_j\right)\right).
\end{equation*}

This is implemented for batched data by the \texttt{log\_prob} method in Listing~\ref{lst:gmm_dist} using JAX's highly performant vectorised mapping capabilities.

Having encapsulated the sampling and log-probability of the Gaussian mixture model in a NumPyro \texttt{Distribution}, we can easily make use of it in our \texttt{model} shown in Listing~\ref{lst:gmm_model} and only need to specify the prior distributions for the parameters of the model. Following \cite{jalko2016differentially}, we use a Dirichlet distribution for $\pi_1, \ldots, \pi_K$ and zero-centered Normal priors for $\bm{\mu}_j$. We assume that each component is spherical, i.e., $\bm{\Sigma}_j = \sigma_j^2\bm{I}$ and use the Inverse Gamma distribution as prior for $\sigma_j$. The code in Listing~\ref{lst:gmm_model} reflects this using the same imperative sampling instructions demonstrated in the earlier examples.

We note that, in principle, it would also have been possible to implement the sampling steps of the GMM steps directly in the \texttt{model} function without the need of subclassing NumPyro's \texttt{Distribution} class. However, this would require learning the values of the latent variables $z_i$ for each data record during inference, which presents a problem for private inference \cite{jalko2016differentially}. The resulting need for a specific implementation of the marginalised log-probability is what makes the GMM an interesting example for the flexibility and expressiveness of NumPyro-based models for privacy-preserving probabilistic programming in \dppp. Note that, compared to the implementation of the experiment in \cite{jalko2016differentially}, we did not have to concern our implementation with effects of reparametrisation on gradients and other details of DP-VI but focus on providing a straightforward implementation of the model.

\begin{lstlisting}[
    float=h, language=Python, label=lst:gmm_dist,
    caption={Implementation of a Gaussian mixture model distribution in NumPyro with log-likelihood marginalised over the latent component assignments.}
]
class GaussianMixtureModel(Distribution):

    def __init__(self, mixture_probabilities, mixture_locs, mixture_scales):
        self._pis = mixture_probabilities
        self._locs = mixture_locs
        self._scales = mixture_scales
        
        batch_shape = ()
        event_shape = self._locs.shape[1:]
        super().__init__(
            batch_shape, event_shape)

    def sample(self, rng_key, sample_shape=()):
        zs_rng, xs_rng = \
            jax.random.split(rng_key)
        zs = CategoricalProbs(self._pis)\
            .sample(zs_rng, sample_shape)
        xs = Normal(
            self._locs[zs], self._scales[zs]
        ).sample(xs_rng)
        return xs

    def log_prob(self, value):
        per_component_log_prob = jax.vmap(
            lambda loc, scale: Normal(
                loc, scale
            ).log_prob(value),
            out_axes=-1
        )(self._locs, self._scales)

        log_pis = jnp.log(self._pis)

        # sum log-likelihood contributions
        # from event dimensions
        per_component_log_prob =\
            per_component_log_prob.sum(axis=-2)

        # aggregate over components
        loglik = logsumexp(
            per_component_log_prob + log_pis,
            axis=-1
        )
        return loglik
}
\end{lstlisting}

\begin{lstlisting}[
    float=h, language=Python, label=lst:gmm_model,
    caption={Definition of the model for a Gausian mixture model, using the \texttt{GaussianMixtureModel} distribution class defined in Listing~\ref{lst:gmm_dist}.}
]
def model(xs, N, k=5, d=2):
    pis = sample('pis', Dirichlet(jnp.ones(k)))

    with plate('component_priors', k, k):
        mus = sample('locs',
            MultivariateNormal(
                jnp.zeros((d,)), jnp.eye(d)
            ), sample_shape=(k,)
        )
        sigmas = sample('sigmas',
            InverseGamma(1, 1),
            sample_shape=(k,)
        )

    batch_size = xs.shape[0]
    with plate('batch', N, batch_size):
        sample(
            'xs', GaussianMixtureModel(
                pis, mus, sigmas
            ),
            obs=xs, sample_shape=(batch_size,)
        )
\end{lstlisting}

\end{document}